\documentclass[runningheads]{llncs}
\usepackage[T1]{fontenc}
\usepackage{graphicx}
\usepackage{amsmath}
\usepackage{amssymb}
\usepackage{subcaption}
\usepackage{multicol}
\usepackage{booktabs}
\usepackage{dblfloatfix}

\newcommand{\eg}{e\/.\/g\/.,\/~}

\newcommand{\wrt}{w\/.r\/.t\/.\/~}

\DeclareMathOperator*{\argmin}{arg\,min}

\begin{document}
\title{Concept Extraction for Time Series with ECLAD-ts}
%
%
\author{Antonia Holzapfel\inst{1} \and
Andres Felipe Posada Moreno\inst{1}\and
Sebastian Trimpe \inst{1}}
\authorrunning{Holzapfel et al.}
%
\institute{Institute for Data Science in Mechanical Engineering (DSME), RWTH Aachen University, 
Theaterstraße 35-39, 
52062 Aachen, Germany
\email{holzapfel@dsme.rwth-aachen.de}\\
}
\maketitle              
\begin{abstract}
Convolutional neural networks (CNNs) for time series classification (TSC) are being increasingly used in applications ranging from quality prediction to medical diagnosis.
The black box nature of these models makes understanding their prediction process difficult. 
This issue is crucial because CNNs are prone to learning shortcuts and biases, compromising their robustness and alignment with human expectations.
To assess whether such mechanisms are being used and the associated risk, it is essential to provide model explanations that reflect the inner workings of the model. 
Concept Extraction (CE) methods offer such explanations, but have mostly been developed for the image domain so far, leaving a gap in the time series domain. 
In this work, we present a CE and localization method tailored to the time series domain, based on the ideas of CE methods for images. We propose the novel method ECLAD-ts, which provides post-hoc global explanations based on how the models encode subsets of the input at different levels of abstraction. 
For this, concepts are produced by clustering timestep-wise aggregations of CNN activation maps, and their importance is computed based on their impact 
on the prediction process. 
We evaluate our method on synthetic and natural datasets. 
Furthermore, we assess the advantages and limitations of CE in time series through empirical results.
Our results show that ECLAD-ts effectively explains models by leveraging their internal representations, providing useful insights about their prediction process.
\keywords{Concept Extraction \and Interpretability \and Time Series.}
\end{abstract}
%
%
%

\section{Introduction}

Convolutional neural networks (CNNs) for time series classification (TSC) are currently being used in the industry, \eg for quality prediction~\cite{yuan2024quality}, as well as in more safety critical domains, like medical diagnosis~\cite{cancers15143608}. 
These models are powerful, but they are also black boxes. 
Therefore, explanations are necessary  to allow developers, users, and stakeholders to understand the operations of models and whether they are aligned with human expectations. 
With such knowledge, the involved agents can improve models, or
choose the `correct' models, upon which they can build an adequate level of trust.
%
A `correct' model is a model that uses information that is relevant for its task, in accordance with human know-how, and avoids the use of harmful biases or spurious correlations present in the data. 
A priori, there is no guarantee that a model will learn to use the desired information.  
Conveniently, Concept Extraction (CE) methods provide such insights
, which would make them useful for time series models.

In the time series domain, global model explanations exist in the form of prototypes~\cite{das2020interpreting}. 
However, these explanations do not directly reflect how the model encodes information, which can be problematic for detecting shortcut learning and other biases. 
Concept testing methods for global explanations have also been applied in time series, \eg to EEG~\cite{madsen2023conceptbased} and bearing fault classification models~\cite{kusters2020conceptual}. 
These methods require the labeling of concepts to test whether the model can distinguish their presence and importance in the inputs, which implies manual effort and expertise in the field of application. 
Additionally, it induces biases with respect to which concepts are tested—the concepts being used by the model do not necessarily coincide with the labeled concepts. 
Similarly, MultiVISION, a Concept Extraction method for time series, has been proposed by~\cite{younis2022multivariate}. 
This method extracts concepts by clustering subsamples corresponding to the receptive field of highly active neurons.
However, MultiVISION's reliance on the receptive field of a network makes it unusable for some deep architectures. 
For example, the receptive field of an InceptionTime network with 10 inception blocks is 343. 
This means that for a network with an input size smaller than 343, the concept patches can be as big as the input, turning the `concept clustering' into input clustering.  
On the other hand, methods for automatic concept extraction exist in the image domain~\cite{kim2018interpretability,kamakshi2021pace,posada-moreno2023eclad}  
and they have shown promising results,
but have not been applied to time series yet. 
A transfer to the time series domain requires considering domain-specific characteristics, such as the dimensionality and channel-specific information. 

In this work, we propose ECLAD-ts, an algorithm that produces post-hoc global explanations through automatic concept extraction and localization. 
Our method is based on ECLAD~\cite{posada-moreno2023eclad,posada-moreno2023scalable} from the image domain, because it provides granular explanations by taking into account the equivariance property of CNNs, which enables not only extraction, but also localization of concepts. 
We introduce mechanisms that make ECLAD suitable for time series. 
ECLAD-ts produces concepts by clustering Local Aggregated Descriptors (LADs), which are timestep-wise aggregations of CNN activation maps. 
Additionally, ECLAD-ts computes an importance score for each concept based on its impact on the model predictions according to the model gradients. 
We test the method on synthetic and natural datasets, and compare it to other CE methods from the time series and the image domain, like vanilla ECLAD, ConceptShap, and MultiVISION. 
Furthermore, we explore the benefits and limitations of such methods in time series. 

We show that time series models encode patterns in their latent space distinctively and thus, concepts can be extracted with ECLAD-ts using the inner representations of the model. 
The extracted concepts 
provide useful insights on the prediction process of the model.

To summarize, the main contributions in this work are:
\begin{enumerate}
    \item ECLAD-ts, a novel algorithm for CE and localization in the time series domain, which captures temporal and channel-wise information.
    \item An importance scoring mechanism that quantitatively assesses the relevance of a concept considering channel-wise gradient information.
    \item Experiments in both synthetic and natural time series, with empirical results that compare several CE methods in time series and highlight their advantages and limitations of CE methods.
\end{enumerate}

\section{Related Work}


The goal of this paper is to enhance the interpretability of deep TSC models through post-hoc explanation, allowing for a better understanding of their alignment with human expectations. 
This research falls under the domain of explainable artificial intelligence (XAI), specifically, post-hoc explainability methods \cite{burkart2021survey,das2020opportunities}. 
Post-hoc methods analyse an existing model after training and provide insights either on single predictions (local) or on the model's general prediction process (global). 
Examples of local explanation methods include saliency maps such as SHAP \cite{lundberg2017unified} and Grad-CAM \cite{selvaraju2017gradcam}, which provide sensitivity scores to quantify the influence of each feature of a datapoint on the prediction. 
In comparison, global explanation methods such as concept extraction \cite{ghorbani2019automatic,kamakshi2021pace,posada-moreno2023scalable,posada-moreno2023eclad} and prototypes \cite{das2020interpreting} analyse a model within the context of a complete dataset, providing explanations about its overall decision-making process.
This is, the insights of global explanations can be extrapolated to new predictions, providing a general rationale of what a model considers in its prediction process.

Our work is related to Concept Extraction methods, which are post-hoc global explanations. 
Concept-based methods explain the decision-making process of a model through the patterns it learns to identify.
These patterns, referred to as concepts, represent human-understandable features that the model has learned to distinguish \cite{kim2018interpretability}. 
Earlier works tested whether a specific concept is present in the model’s internal representations by comparing how sets of manually labeled instances are encoded within the latent space of models.
Representative concept testing approaches include TCAV \cite{kim2018interpretability} and CAR \cite{crabbe2022concept}. 
In contrast, Concept Extraction focuses on automatically identifying and isolating these concepts. 
Methods such as ACE \cite{ghorbani2019automatic}, ConceptShap \cite{yeh2020completenessaware}, PACE \cite{kamakshi2021pace}, SPACE \cite{posada-moreno2023scalepreserving}, and ECLAD \cite{posada-moreno2023scalable,posada-moreno2023eclad} have been primarily proposed for the image domain, providing significant insights, but have not been transferred to the time series domain yet.
The main reason being the particularities of time series data, such as the dimensionality, or channels having disentangled information, which results challenging for current CE methods.
In this context, we propose extending and transferring the CE method ECLAD for analysing CNNs used in TSC.

In the domain of time series, five concept-based approaches have been proposed. 
The first two approaches perform concept testing with TCAV \cite{kim2018interpretability} for concepts in EEG \cite{madsen2023conceptbased} and bearing fault classification \cite{decker2023explaining}. 
In these approaches, sets of samples are manually annotated or modified to contain specific concepts, to test whether the model can distinguish their presence in an input. 
Similarly, the third approach relates concepts to the presence of specific frequencies by creating negative samples through handcrafted filters applied to the time series \cite{kusters2020conceptual}.
The fourth approach relies on preprocessing data through an autoencoder and analysing the impact of the autoencoder's latent dimensions on the classification of the reconstructed signal \cite{obermair2023example}. 
Lastly, the fifth approach performs concept extraction by clustering slices of the inputs that correspond to the receptive field of highly active neurons.
The current approaches either (1) rely on manual annotation, and creation of concept samples, (2) are directly unrelated to the latent space of models, or (3) rely on the receptive field of NN layers for the extraction of concepts. 
The first case requires significant human involvement for manual annotation. 
The second case ties concepts to features directly extracted from the data, which are not necessarily the features learned by models, especially in cases of shortcut learning \cite{geirhos2020shortcut}, defeating the purpose of these explanations. 
The third case is not usable for deep models, since their receptive field can become too large to produce meaningful concepts.
For example, a 1D DenseNet121 has a large enough receptive field such that clustering is performed over the whole input. 
In our work, we propose a CE method to automatically extract concepts based on how models encode time spans of the input through different levels of abstraction, without relying on human annotations nor on the receptive field of the NN. 
Our approach provides concept extraction and localization capabilities directly related to how the model encodes input information and uses it for its prediction process.

\section{Methods} \label{sec:methods}
In this section, we explain the ECLAD-ts algorithm and our experimental setup. 
This includes the validation of our method using synthetic datasets and implementation details.  

\subsection{ECLAD-ts}
    In this work, we use ECLAD~\cite{posada-moreno2023eclad,posada-moreno2023scalable} from the image domain as a base framework to develop our method for performing CE in TSC models. 
    ECLAD is an algorithm that provides explanations based on three key steps: The encoding of the latent space of neural networks, the mining of patterns and the assessment of how relevant they are for the prediction process of the model. %
    ECLAD-ts has two main differences with respect to ECLAD:
    \begin{enumerate}
        \item The latent space representation is modified to be compatible with time series data.
        \item The importance score is modified to account for the importance of channel-specific information and for concepts that affect all classes similarly, which are pervasive in time series models.
    \end{enumerate}
    Below, we describe in detail each step of ECLAD-ts. 
    
    \subsubsection{Encoding of the latent space} %
        In the first step, ECLAD-ts uses the notion of \textbf{Local Aggregated Descriptors (LADs)}, which are timestep-wise descriptors of how models encode a region at different levels of abstraction. 
        These are obtained by aggregating the activation maps of multiple layers $l$ of a CNN model. 
        
        In the case of TSC with $n_k$ classes, a typical CNN maps the inputs of the model $x_{i} \in \mathbb{R}^{w \times d}$, to an output $y \in \mathbb{R}^{n_k}$ with a function $f: x \mapsto y$, where $w$, and $\textrm{ch}$ are the length and channels of the input. 
        The function $f$ is a composition of multiple functions ordered in layers, where the model encodes the information of the input into the latent spaces of each layer. 
        An activation map $a_l = f_l(x)$ belonging to the latent space $\mathbb{R}^{\textrm{ch}_l,w_l}$ of layer $l$ is computed by a partial evaluation of the model until said layer. 
        The dimensions of $a_l$ depend on the type of layer evaluated. 
        
        We aggregate the activation maps of a predefined set of $n_l$ layers $L = \{l_1, \dots, l_{n_l}\}$, upscaled with linear interpolation ($f_U$) to the dimensions of $x_i$. This produces a descriptor $d_{x_i} = [ f_U(f_{l_1}(x_i)) \dots f_U(f_{l_{n_l}}(x_i))] \in \mathbb{R}^{w \times \textrm{ch}^*}$, where $\textrm{ch}^*$ is the sum of the number of units in all layers in $L$. A LAD refers to each timestep $d_{x_i}(b)\in \mathbb{R}^{ 1\times \textrm{ch}^* }$ of $d_{x_i}$, where $b$ denotes its position along the width of $d_{x_i}$.
        LADs contain information about the model encoding of timesteps at different levels of abstraction. 
        Here, we effectively reduced the dimensionality of the LADs introduced in ECLAD.
        
    \subsubsection{Mining of patterns and visualization}
        In the second step of ECLAD-ts, we separate the latent space of the model into clusters of different patterns using LADs. 
        For this, we apply a \textbf{mini-batch k-means algorithm} to the LADS extracted from a set of inputs and obtain a set $\Gamma = \{ \gamma_{c_1}, \dots, \gamma_{c_{n_c}} \}$ of centroids $\gamma_{c_j} \in \mathbb{R}^{1\times \textrm{ch}^* }$ defining the \textbf{concepts}. The centroids represent similarly encoded time subsequences. 

        
        For a human-understandable visualization, the concepts $c_j$ can be located in an input $x_{i}$ by creating a \textbf{mask} $m_{x_{i}}^{c_j}\in \mathbb{R}^{ w \times 1}$ that analyses the LAD at each timestep, and assesses whether they belong to a cluster $\gamma_{c_j}$ as in
         \begin{equation}
              m_{x_{i}}^{c_j}(b) = \begin{cases} 
               1 & \argmin_{c_q}(||d_{x_{i}}(b) - \gamma_{c_q}||_2) = c_j \\ 
               0 & \mathrm{otherwise}.
            \end{cases}
         \end{equation}
        Each of these masks has only one channel. 
        For both visualization and importance score computation, the masks are expanded to have the same number of channels than the input.
        This produces $\textrm{ch}$ expanded masks for each concept, each containing zeros at all channels except the $p$-th, $p\in\{1,\dots, \textrm{ch}\}$, which contains the original mask $ m_{x_{i}}^{c_j}$.

    \subsubsection{Relevance computation}
    A key contribution of our work is the way we calculate the relevance of each concept.
    For the last step of ECLAD-ts, the \textbf{importance score (IS)} of a concept is determined, which quantifies the relevance of its related visual cues towards the prediction of the analysed model. 
    Building on ECLAD's, we propose a new metric that is compatible with multilabel classification and can discriminate the importance of a concept over multiple channels for multivariate timeseries. 
    The reason for this is that, unlike images, timeseries often have channels encoding different types of information (\eg channels that correspond to different kinds of sensor) and thus contain information that can vary in importance for the same timestep.
    Therefore, it can be necessary to know which channel is important for a concept explanation.
    
    To determine the IS, we first compute the the instance-wise sensitivity $R_{x_{i}}^{c_j}$ of the timeseries regions belonging to a concept, 
    \begin{align}
        \label{eq:img_r}
        R_{x_{i}}^{c_j} = \nabla_x g(f(x_{i})) \odot  m_{x_{i}}^{c_j},  & \textrm{\phantom{bla}} R_{x_{i}}^{c_j} \in \mathbb{R}^{\textrm{ch} \times w}, 
    \end{align}

    where $\odot$ denotes the element-wise product between matrices and $g(y)$ is a wrapper of the model $f$. 
    The wrapper \(g(y)\) is defined as
    \[
    g(y) = \|y\cdot \boldsymbol{1}^T - \boldsymbol{1}\cdot y^T\|_2,
    \]
    for multiclass classification, where \(y\in \mathbb{R}^{n_k}\) is the output of \(f\), \(n_k\) is the number of classes, and \(\boldsymbol{1}\) is the vector of ones of the same size as \(y\).

    The timestep-wise sensitivity for a channel $\textrm{ch}_p$ is computed as 
    \begin{equation} \label{eq:sum_IS}
        r_{x_{i}, \textrm{ch}_p}^{c_j} = \sum_{b \in w} R_{x_{i}, \textrm{ch}_p}^{c_j}(b)
    \end{equation}
    where  $R_{x_{i}, \textrm{ch}_p}^{c_j}(b) \in \mathbb{R}^{1 \times 1}$ refers to each timestep of $R_{x_{i}}^{c_j}$ at channel $\textrm{ch}_p$.        
    
    This step is modified from the metric used in ECLAD by replacing the 1-norm in Eq.~4 of \cite{posada-moreno2023scalable} by the sum over the timesteps. 
    The rationale behind this is that concepts having a negative gradient in Eq.~\ref{eq:img_r} are concepts that reduce the distance between the logits. 
    This means that they are concepts that make the model `less sure' about its output, which is the opposite of what concepts with a positive gradient do.
    Thus, it is important to retain the sign of the gradient for computing the importance score, which is achieved by our modification in Eq.~\ref{eq:sum_IS}.  
    While this characteristic of the importance score might not have been noticeable in the image domain, it has a visible effect in the time series domain. 

    Using our IS metric, a concept with an importance score close to 1 is a concept that is useful for distinguishing between the classes, while a concept with an importance score close to -1 is a concept that puts in question the predicted class and, in a way, represents the features that make the model unsure of its output. 
    Consequently, a concept with a score close to 0 is `unimportant'
    
    The IS computation ends by aggregating $r_{x_{i}}^{c_j}$ over the timeseries in the dataset $E$ and scaling the mean relevance of each concept to obtain the final IS per channel 
    \begin{align}
        I^{\textrm{ch}_p}_{c_j}= \frac{r^{\hat{c}_j}}{\max_{c_j} | r^{\hat{c}_j} |} \, & \textrm{, where} & r^{\hat{c}_j}= \frac{1}{n_{c_j}} \sum_{x_i \in E} r^{c_j}_{x_i} &
    \end{align}
    and $n_{c_j}$ is the number of datapoints containing a concept. The resulting $I^{\textrm{ch}_p}_{c_j}$ is the importance of a concept $c_j$ in channel $\textrm{ch}_p$.

    With these three steps, ECLAD-ts can extract patterns that are meaningful for the model and can be represented in human-understandable visualizations. 
    These patterns leverage the equivariance properties of the models. 

\subsection{Validating CE}
Testing CE methods is challenging, because we do not have access to the `ground truth' of the patterns learnt by a CNN. %
To validate CE methods, we can instead generate synthetic datasets using primitive concepts (\eg a local structure) that are important for the labels by design.
We understand primitives to be patterns that have associated annotations as binary masks, denoting their position. 

If the synthetic dataset is designed correctly, a good model must use at least a subset of the the patterns in the primitives for its task. 
Thus, in an ideal case, the concepts related to primitives would have high importance scores, and the concepts unrelated to them would have low importance scores. 
Considering this, the primitives can serve as a surrogate for the ground truth of an accurate model. %
It is important to note that the primitives cannot be considered equivalent to a ground truth. %
Due to shortcut learning, models can learn a sufficient subset of features and still achieve perfect accuracy. 
This means that for some primitives, the importance score is not going to be close to 1, even though they are designed to be important.  %
Nevertheless, it stands to reason that for a model with $N$ primitives and $N$ classes, at least $N-1$ concepts related to the primitives are needed for proper classification. 
Given this, the primitives remain useful to test a model's concepts. 

For the objective validation of CE methods, we consider particular aspects. 
First, given that the primitives serve as a proxy ground truth, the extracted concepts should be temporally consistent with these ground truth primitives—i.e., their localization masks must align closely with those of the primitives. 
Second, the importance assigned to each concept should reflect its association with critical primitives; 
concepts closely associated to important primitives should receive high importance scores, whereas those with weaker associations should be deemed less important. 
This dual criterion ensures that the concepts not only mirror the ground truth in their temporal location but also capture the intended relevance of the underlying features.

To quantify the precision of a CE method, as in~\cite{posada-moreno2023eclad}, we compute the two-way distance $\mathrm{DST}_{p_0,c_j}$ to compare the masks of concepts learned by the model with those of the primitives.
We associate each concept to its closest primitives, respecting a minimum threshold (here, 20\% of the maximum possible distance between concept masks and primitives).
As a result, we clarify which concept represents which primitive from the ground truth, labeling as aligned concepts if they are closely related to important primitives.
Furthermore, we apply the Representation Correctness (RC) and Importance Correctness (IC) metrics form ~\cite{posada-moreno2023eclad} to quantify the alignment of the extracted concepts with the visual cues and intended importance of the features of the dataset.
\textit{Representation correctness} is defined as the average of the negative association distance $\mathrm{DST}_{p_0,c_j}$ of all aligned concepts extracted from the examined models.
Whenever no concept alignments are found, the Representation correctness score is set to 40\% of the maximum penalty for visualization purposes. 
\textit{Importance correctness} is the mean importance of all aligned concepts, minus the mean importance of all unaligned concepts, normalized by the maximum importance of all concepts. 
The idea behind these metrics is that aligned concepts should be well represented (RC close to zero) and scored as important (high IC), whereas unaligned concepts should be scored as unimportant. 
%

\begin{figure}[!b]
    \centering
    \includegraphics[width=\linewidth]{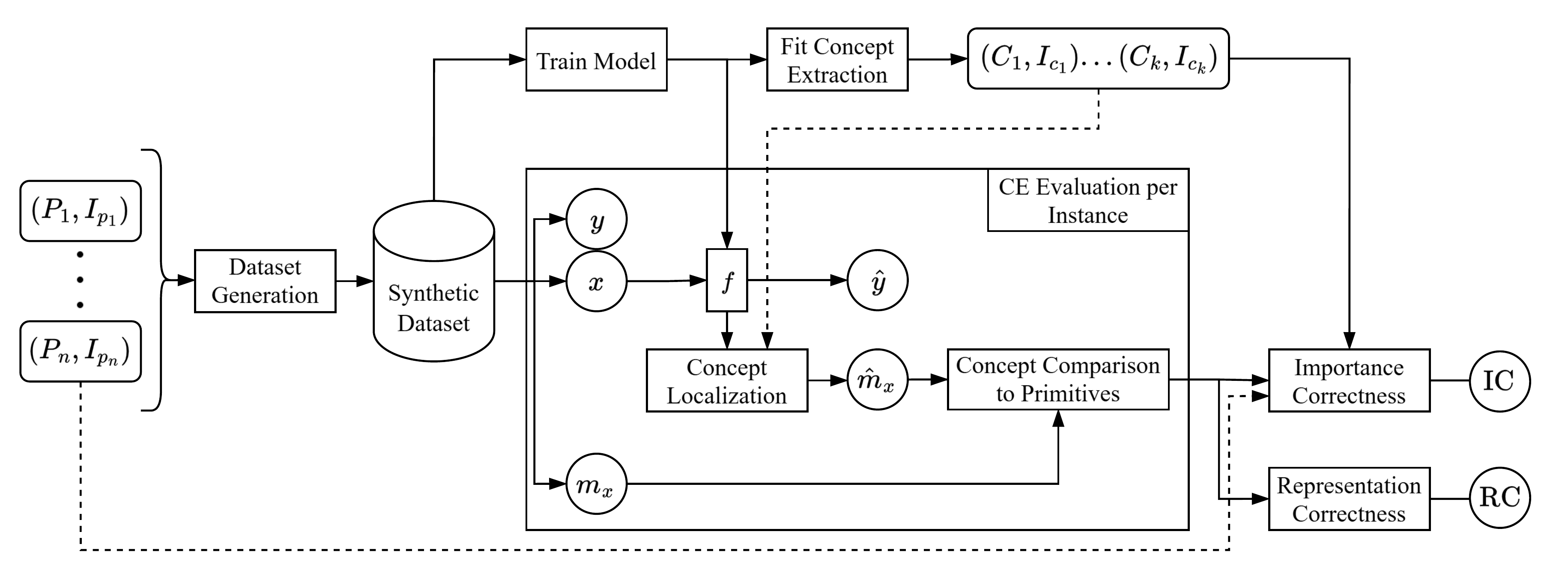}
    \caption{Validation Pipeline for Concept Extraction. (1) Primitives definition. (2) Synthetic dataset generation. (3) Models are trained on the synthetic dataset and concept extraction is performed. (4) Extracted concepts are localized and compared with the ground truth primitives to establish associations. (5) Evaluation metrics for representation and importance correctness are computed.
    }
    \label{fig:validationpipe}
\end{figure}%

The validation pipeline CE methods is illustrated in Fig.~\ref{fig:validationpipe}, the step by step process is: 
(1) Define a set of primitive concepts along with their associated importance scores. 
(2) Generate a synthetic dataset using these primitives, where each data point is a tuple $(x, y, m_x)$—with $x$ as the input, $y$ as the output, and $m_x$ as the primitives mask. 
(3) Train several models $f$ on the synthetic dataset, then fit a concept extraction method to obtain a set of concepts and their importance scores. 
(4) For each instance, localize each concept and obtain their masks to associate each concept with its closest primitive. 
(5) Finally, compute the evaluation metrics—Representation Correctness (RC) and Importance Correctness (IC)—to assess how well the extracted concepts capture the intended visual cues and importance of the primitives.

For the purpose of this work, we generated three synthetic datasets. 
The syntheticL2 and syntheticL4 datasets each have local features as primitives and a single channel.
The syntheticLm dataset also has local features as primitives, but two channels. 
The purpose is to test if CE methods can extract meaningful concepts that are close to the primitives in position and importance. 
By `meaningful concept' we refer to patterns in the latent space that translate to recognizable patterns in the input space. 
During our validation, we evaluate both objectively (through the introduced metrics), as well as qualitatively, if the concepts provide further insights in the models' prediction processes. 

\subsubsection{syntheticL2 and syntheticL4}
\begin{figure}[!h]
    \centering
    \begin{minipage}{.23\linewidth}
    $C_0$
        \begin{subfigure}[t]{\linewidth}
            \includegraphics[width=\linewidth, height =0.5\linewidth]{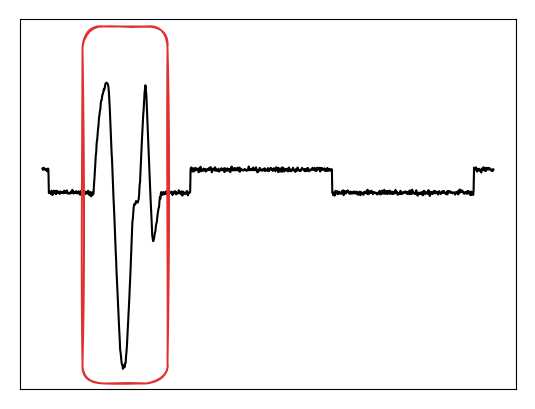}
        \end{subfigure}
    \end{minipage}
    \begin{minipage}{.23\linewidth}
    $C_1$
        \begin{subfigure}[t]{\linewidth}
            \includegraphics[width=\linewidth, height =0.5\linewidth]{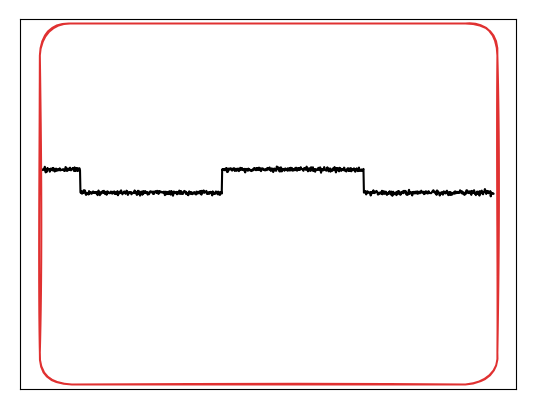}
        \end{subfigure}
    \end{minipage}
    \phantom{}
    \begin{minipage}{.23\linewidth}
    $C_0$
        \begin{subfigure}[t]{\linewidth}
            \includegraphics[width=\linewidth, height =0.5\linewidth]{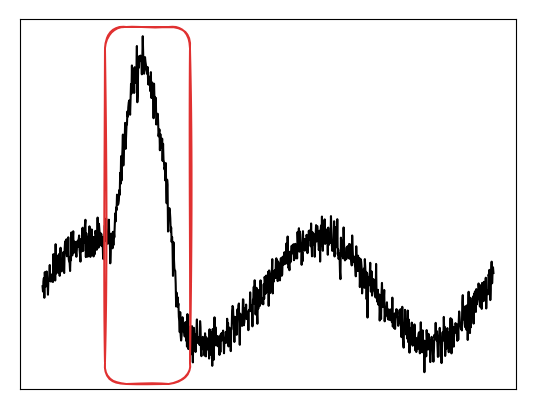}
        \end{subfigure}
    \end{minipage}
    \begin{minipage}{.23\linewidth}
    $C_1$
        \begin{subfigure}[t]{\linewidth}
            \includegraphics[width=\linewidth, height =0.5\linewidth]{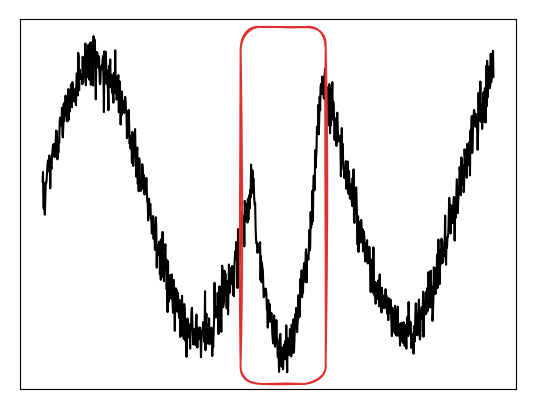}
        \end{subfigure}
        \end{minipage}
    \caption{Left: Examples of class 0 and 1 of syntheticL2 dataset. Right: Examples of class 0 and 1 of syntheticL4 dataset. The red boxes signalize the primitives used for generating the samples of each class.}
    \label{fig:primitives}
\end{figure}

The syntheticL2 and syntheticL4 datasets have two classes that are generated using primitives. 
SyntheticL2 is generated using primitive $p_0$ in class 0, and no `no $p_0$' in class 1. 
Both classes have a background consisting of square wave and white noise.
Uninterrupted background is considered to be a second primitive ($p_1$), as it can be used by the model instead of $p_0$. 
SyntheticL4 is generated using two primitives, $p_0$ being in class 0 and $p_1$ being in class 1.  
All primitives for these datasets are designed to be important and will thus make close concepts eligible for the alignment criterion in the RC and IC scores. 
Both classes have a background consisting of a sine wave and Gaussian noise.
The classes are balanced for both synthetic datasets. 
Examples of these datasets are depicted in Fig.~\ref{fig:primitives}. 

\subsubsection{syntheticLm}
\begin{figure}[!h]
    \centering
    \begin{minipage}{.32\linewidth}
    $C_0$
        \begin{subfigure}[t]{\linewidth}
            \includegraphics[width=\linewidth]{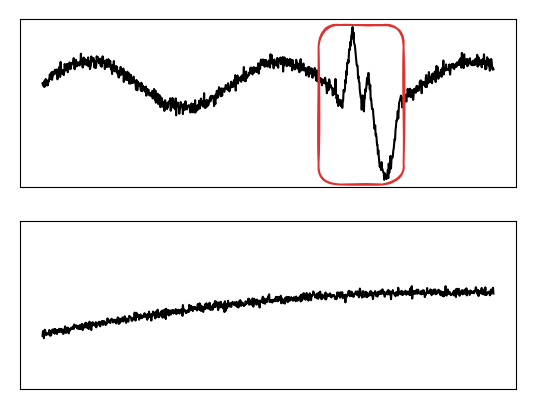}
        \end{subfigure}
    \end{minipage}
    \begin{minipage}{.32\linewidth}
    $C_1$
        \begin{subfigure}[t]{\linewidth}
            \includegraphics[width=\linewidth]{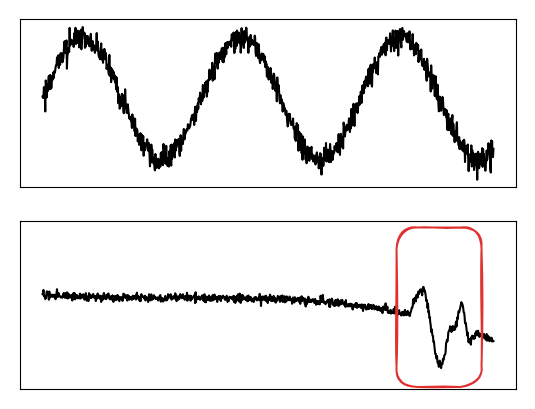}
        \end{subfigure}
    \end{minipage}
    \begin{minipage}{.32\linewidth}
    $C_2$
        \begin{subfigure}[t]{\linewidth}
            \includegraphics[width=\linewidth]{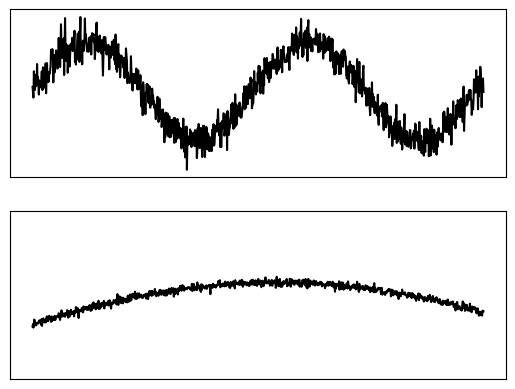}
        \end{subfigure}
    \end{minipage}
    \caption{Examples of classes 0, 1, and 2 of syntheticLm dataset. The red boxes signalize the primitives used for generating the samples of each class.}
    \label{fig:primitives2}
\end{figure}

The syntheticLm dataset has three classes, generated from two primitives. 
Class 0 has $p_0$ at channel 0, while class 1 has $p_1$ present at channel 1, and class 3 has no primitives present. 
Each case happens with 33\% probability, and random noise is added to all samples. 
Additionally, as background, channel 0 has a sine wave and channel 1 has a random polynomial of degree up to 5.
Examples of each of the classes can be seen in Fig.~\ref{fig:primitives2}.

\subsection{Implementation details}
In this work, an experimental run consisted of training a model with a dataset and a specific random seed, and then applying CE as described below. 
Our implementation will be made publicly available upon publication of this work.

\subsubsection{Data} The datasets used in this paper are the syntheticL$n_\textrm{DS}$, syntheticLm, 
the GunPoint dataset from the UCR archive~\cite{UCRArchive2018} and the Production Press Sensor Data (P2S) dataset from ~\cite{kraus2024right}. 
All datasets are used for  multiclass classification.  
Several data augmentations including resizing, time warping and Gaussian noise were applied to the data. 
These datasets were chosen for a general validation of the CE methods: the three synthetic datasets enable an objective validation of the CE methods in controlled scenarios with multiple ground truth primitives, GunPoint provides a straightforward test in a different real-world domain, and P2S presents a more complex industrial use case where the conditions for classification and the related more challenging features.

\subsubsection{Models} 
To evaluate the robustness and generalization capability of our concept extraction methods, we trained three different 1D CNN architectures. 
The CNN architectures were InceptionTime~\cite{ismail2020inceptiontime} with 10 inception blocks, resnet18~\cite{he2016deep} and DenseNet121~\cite{huang2017densely}. 
Furthermore, we repeated the experiments with 10 random seeds (seeds 0 to 9), allowing us to account for the stochastic nature of both model training and the explanation (concept extraction) process. 
Each model was trained until convergence, using early stopping with a patience of 15 epochs. 
The training was performed using the Adam optimizer with weight decay set to $0.01$ and an initial learning rate of $1e^{-5}$. 
We used a ReduceLROnPlateau scheduler with a factor of 0.1 based on the NLL of the models. 
The data was split into 0.8 for training and 0.2 for validation. The same split was used for extracting the concepts with ECLAD-ts and visualizing them.

\subsubsection{Concept Extraction} 
All CE methods were implemented using Pytorch.
At least 2560 samples were used for CE for each model, and the same seed was used as for the model training. 
We tested a range of hyperparameter values for the number of concepts, namely $\{3,5,10,15,20\}$, and the final number of concepts shown in the reports was chosen via visual inspection to ensure they are representative of the overall results. 
The layers used for each CE method can be found in Tab.~\ref{tab:layers}. 
They were chosen at regular intervals for ECLAD and ECLAD-ts.
For ConceptShap and MultiVISION, they were chosen at bottlenecks to avoid examining possibly skipped blocks due to residual connections. 
\begin{table}[h!]
    \caption{Layers for CE with each method and model}
    \label{tab:layers}
    \centering
    \begin{tabular}{c|p{9.5cm}}
    \toprule
         CE methods & ECLAD and ECLAD-ts  \\
    \midrule
         InceptionTime10 & `model.inception\_block.inception\_layers.$n_b$.bottleneck', $n_b\in\{6,7,8,9\}$\\
          ResNet18 &`model.layers.$n_b$.1.relu', $n_b \in \{0,1,2,3\}$ \\
          DenseNet121 &`model.features.transition$n_b$.conv', $n_b \in \{0,1,2,3\}$ \\
          &`model.features.denseblock4.block.15.conv2'\\
    \bottomrule
    \end{tabular}
        \begin{tabular}{c|p{9.5cm}}
    \toprule
         CE methods & ConceptShap and MultiVISION  \\
    \midrule
         InceptionTime10 & `model.inception\_block.inception\_layers.9.bottleneck'\\
          ResNet18 &`model.layers.2.1.relu'\\
          DenseNet121 & `model.features.transition3.conv'\\
    \bottomrule
    \end{tabular}
\end{table}

Additionally, for ConceptShap, Shapley values were computed using a Monte Carlo approximation as described in the original implementation. 
Specifically, we performed 20 iterations of Monte Carlo estimation with hyperparameters set to $\lambda_{\textrm{CS},1}=0.0001$, $\lambda_{\textrm{CS},2}=0.1$, and $\beta_\textrm{CS}=0.2$. The $\lambda$ terms were modifited for convergence, while $\beta_\textrm{CS}$ followed the original recommendation of the method.

For MultiVISION, we set the threshold for neuron activation extraction to the 0.99 quantile of the examined layer. K-means clustering was employed for grouping activations, leveraging its partial fit capability for scalability and to enhance the stability of the resulting clusters. 
Additionally, we adapted the representativeness metric as an importance score, defined as the frequency of a concept per class, where the maximum frequency across classes is normalized by the overall maximum frequency.


\section{Results}

In this section, we show the results of CE with ECLAD-ts and the three compared benchmarks (ECLAD, ConceptShap, and MultiVISION) on three CNN model architectures trained on synthetic and natural datasets, performed as described in the methods section. %
We must highlight that the concept mining of ECLAD-ts is identical to that of ECLAD (which was adapted from the original image ECLAD directly to time series), but it uses our modified channel-wise visualization and our modified IS. 
Thus, the RC of ECLAD and ECLAD-ts should be very similar for univariate datasets. 
We first analyze synthetic datasets—where ground truth primitives are known—to demonstrate how ECLAD‑ts accurately localizes and scores key features. 
We then validate these findings on natural datasets, highlighting our method’s ability to adapt to real‑world data.
In each case, we show a representative example for CE and describe the obtained insights. %
The key takeaways of this result section are that: 
\begin{enumerate}
    \item Useful patterns are encoded distinctively within the latent space of models.
    \item For synthetic datasets, concepts are closely related to primitives and a subset of them are scored with the intended importance. This demonstrates that ECLAD-ts works as intended.
    \item ECLAD-ts allows for understanding the prediction process and detecting shortcut learning in models. 
    \item ECLAD-ts is able to localize concepts not only in time, but also channel-wise. 
    \item ECLAD-ts outperforms the benchmarks in terms of representation and importance correctness.  
\end{enumerate}

\subsubsection{Evaluation on Synthetic Datasets}
Experiments on synthetic datasets were conducted because the underlying primitives are predefined and easily interpretable, allowing for controlled and simplified experiments where models learn these primitives as proxy ground truth of concepts. 
Visualization of the extracted concepts provides a crucial sanity check for the concept extraction methods. 
Moreover, the controlled environment of synthetic data enables the computation of objective metrics—namely, representation correctness and importance correctness—that quantitatively evaluate the alignment between extracted concepts and the underlying primitives in terms of localization and relevance.

\begin{figure}[!htb]
    \begin{minipage}{.48\linewidth}
    \centering ECLAD-ts\\
        \begin{subfigure}[t]{\linewidth}
            \includegraphics[width=\linewidth]{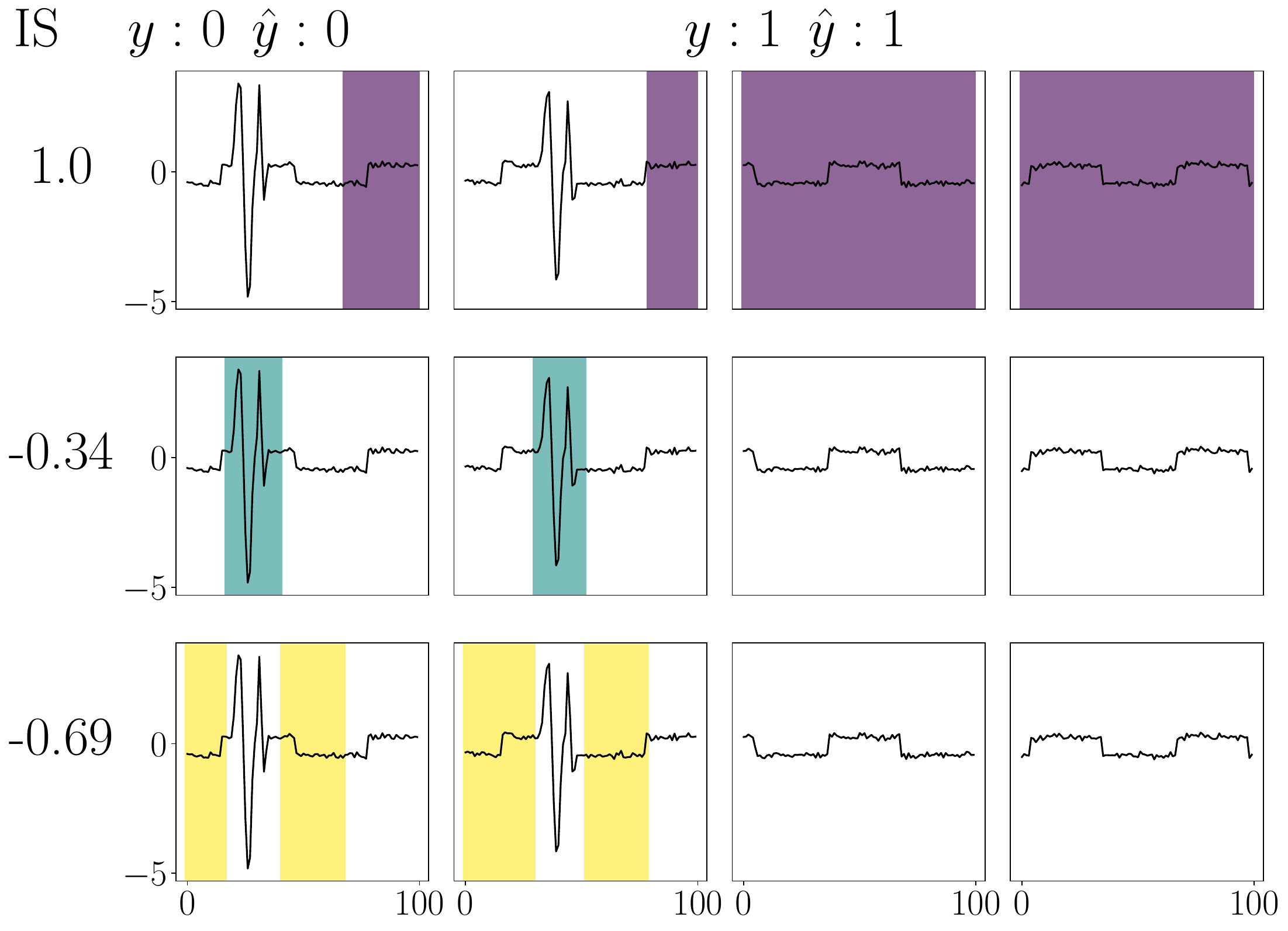}
        \end{subfigure}
    \end{minipage}%
    \begin{minipage}{.48\linewidth}
    \centering ECLAD\\
        \begin{subfigure}[t]{\linewidth}
            \includegraphics[width=\linewidth]{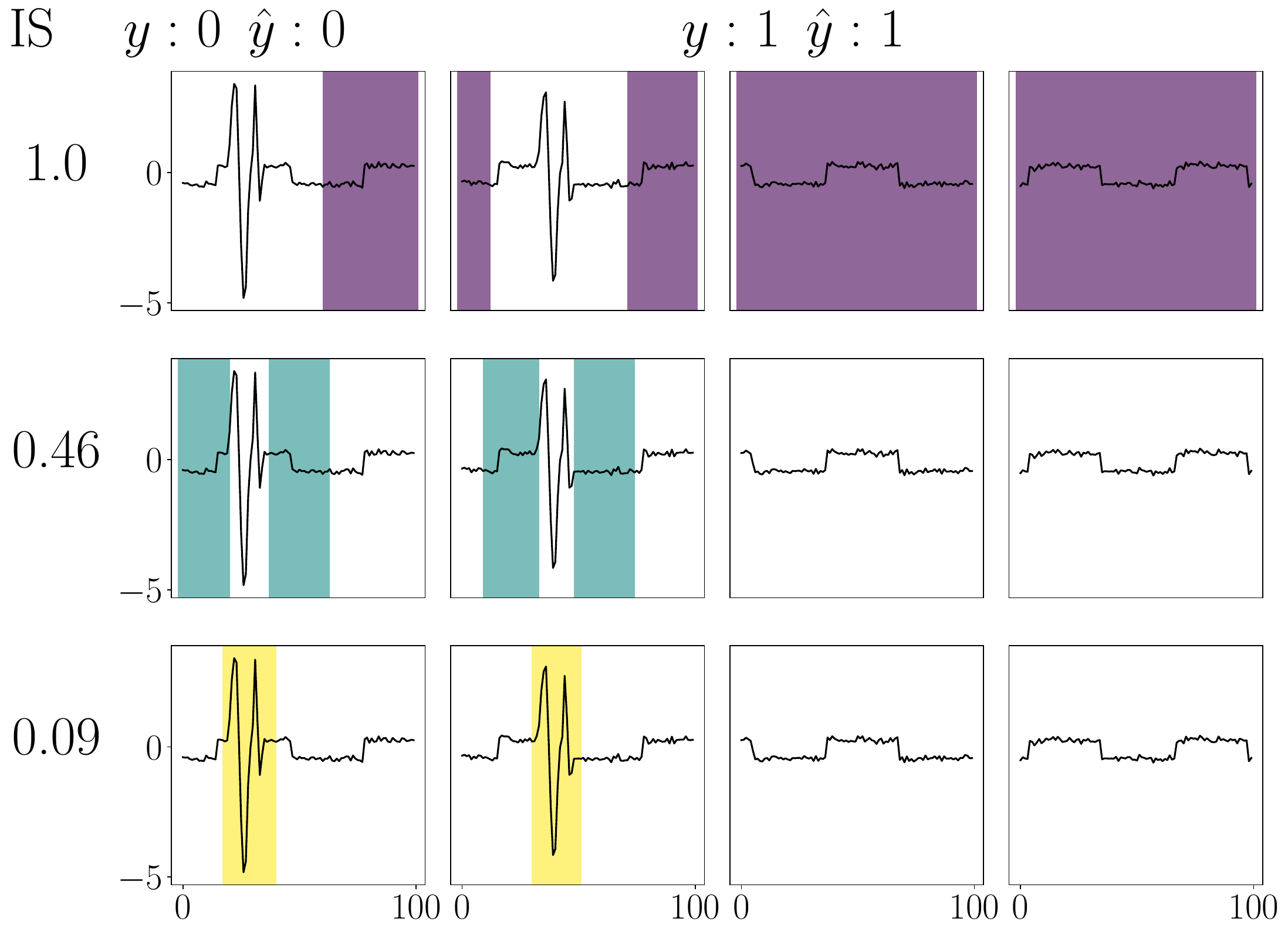}
        \end{subfigure}
    \end{minipage}
    \begin{minipage}{.48\linewidth}
    \centering ConceptShap\\
        \begin{subfigure}[t]{\linewidth}
            \includegraphics[width=\linewidth]{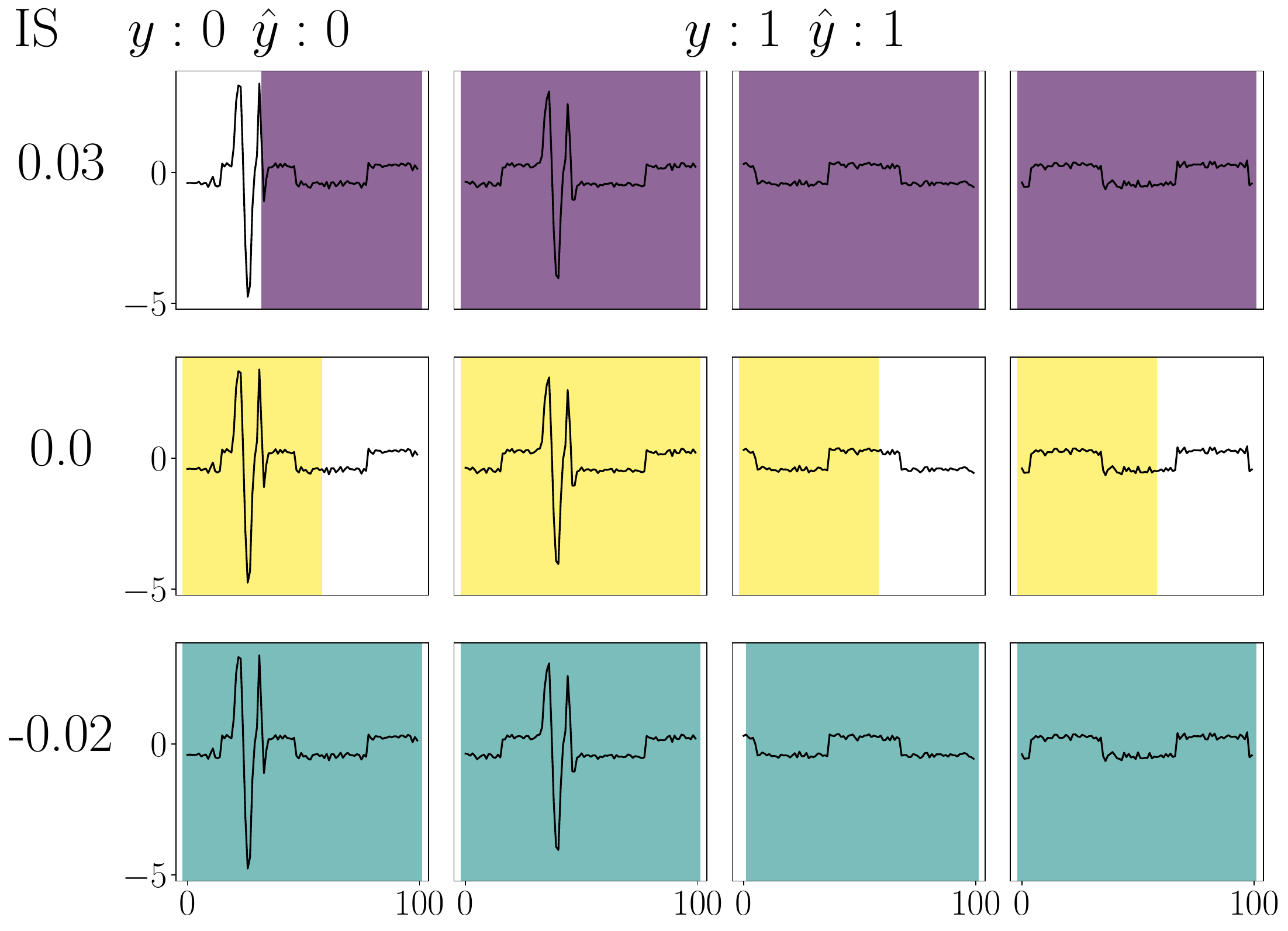}
        \end{subfigure}
    \end{minipage}%
    \begin{minipage}{.48\linewidth}
    \centering MultiVISION\\
        \begin{subfigure}[t]{\linewidth}
            \includegraphics[width=\linewidth]{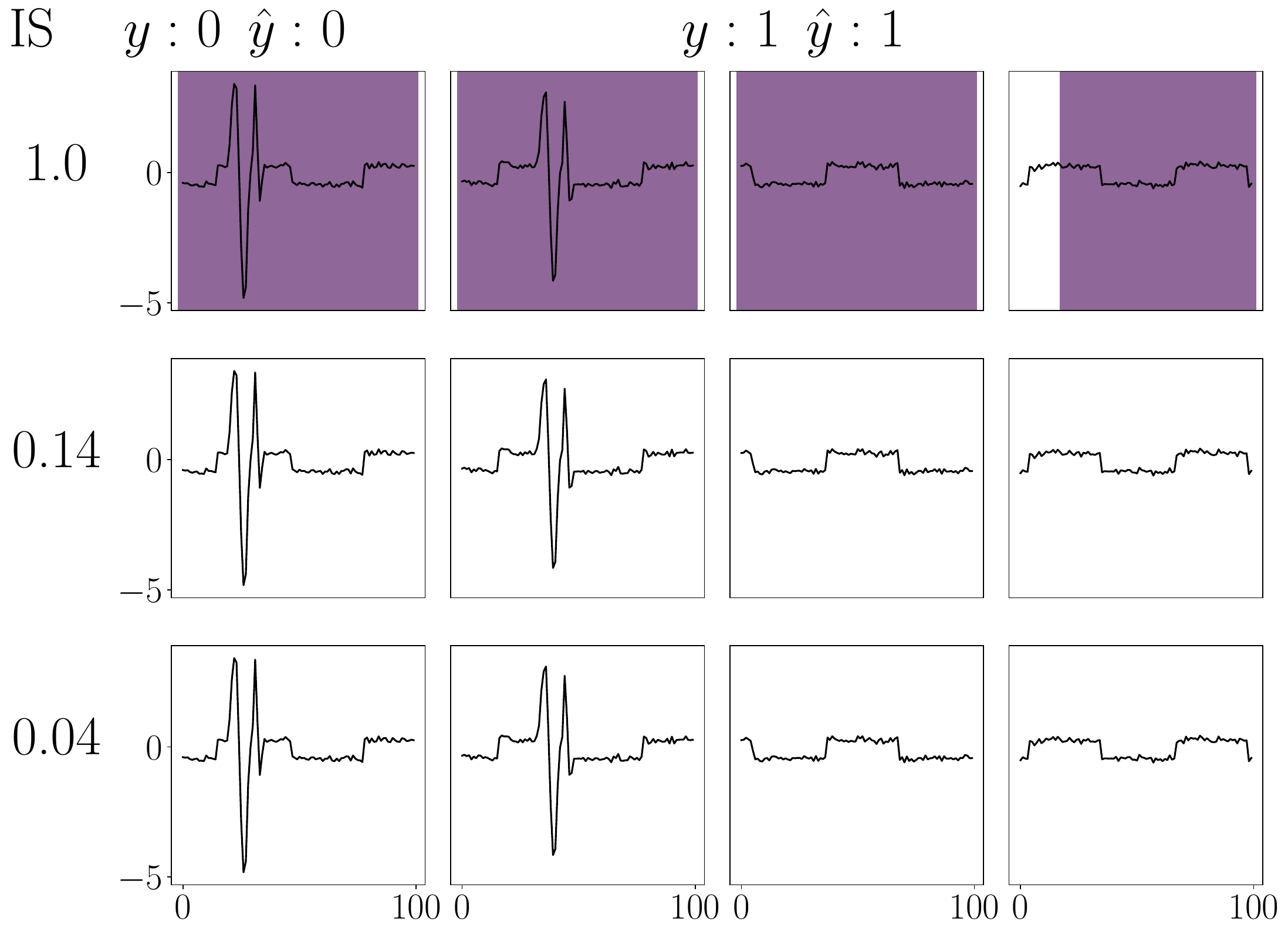}
        \end{subfigure}
    \end{minipage}%
    \caption{
    Concept extraction from ResNet18 on the syntheticL2 dataset is illustrated for four methods. 
    The dataset consists of one channel with a primary primitive \(p_0\) and an uninterrupted background \(p_1\), both serving as discriminative cues. The model (seed=1) achieved a validation accuracy of $100\%$. 
    In each panel, rows denote individual concepts and columns represent instances (with headers showing actual/predicted labels: the first two for class 0 and the latter two for class 1), highlighted regions indicate where concepts appear, and left labels report their importance scores. 
    Notably, ECLAD-ts and ECLAD extract \(p_0\)-related concepts in \(c_1\) and \(c_2\), respectively, while ConceptShap and MultiVISION capture background cues.
    }
    \label{fig:L2_RN}
\end{figure}

The \textbf{syntheticL2} dataset has one channel and a single primitive $p_0$ which is present in half of the instances. 
`Uninterrupted background' is also considered to be a primitive $p_1$, since a model can also discriminate between the classes by (exclusively) detecting the lack of background interruptions. 
Fig.~\ref{fig:L2_RN} shows a representative example of the CE results for the syntheticL2 dataset.
This figure shows the concepts extracted from a ResNet18 model trained on the syntheticL2 dataset, as well as their importance scores. %
We observe that the first and most important concept for all models seems to be related to the background (\(p_1\)). 
Concepts 2 and 3 for both ECLAD-based methods are also distinctively related to primitive $p_0$, but are less important.
For ConceptShap, they also seem to relate to a broad area that can be interpreted as the background, and for MultiVISION, they are empty for the observed samples, but otherwise relate similarly to the background. 

In all cases, the CE methods show that the CNN is using the background as a primary classification cue. 
Nonetheless, the difference between the background and actual features is not present in the concept representation of ConceptShap or MultiVISION. 
In contrast, ECLAD-ts and ECLAD are able to extract concepts relating to both the background and to the primitive $p_0$ of the dataset.
This is, the LADs show that the models encode these features differently, which provides valuable information for understanding the model's behavior. 
Furthermore, according to ECLAD-ts concepts and scores, the  CNN is performing shortcut learning in the sense that it does not rely directly on $p_0$ to make a prediction, but is mostly detecting whether a ``background uninterrupted by a structure (like $p_0$)" is present. 

\begin{figure}[!t]
    \begin{minipage}{.5\linewidth}
        \begin{subfigure}[t]{\linewidth}
            \includegraphics[width=\linewidth]{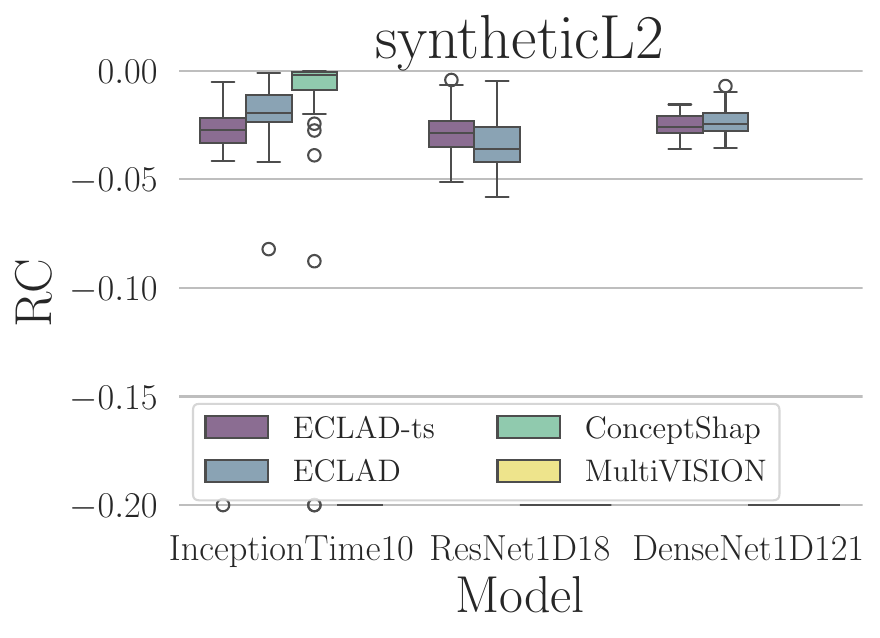}
        \end{subfigure}
    \end{minipage}%
    \begin{minipage}{.5\linewidth}
        \begin{subfigure}[t]{\linewidth}
            \includegraphics[width=\linewidth]{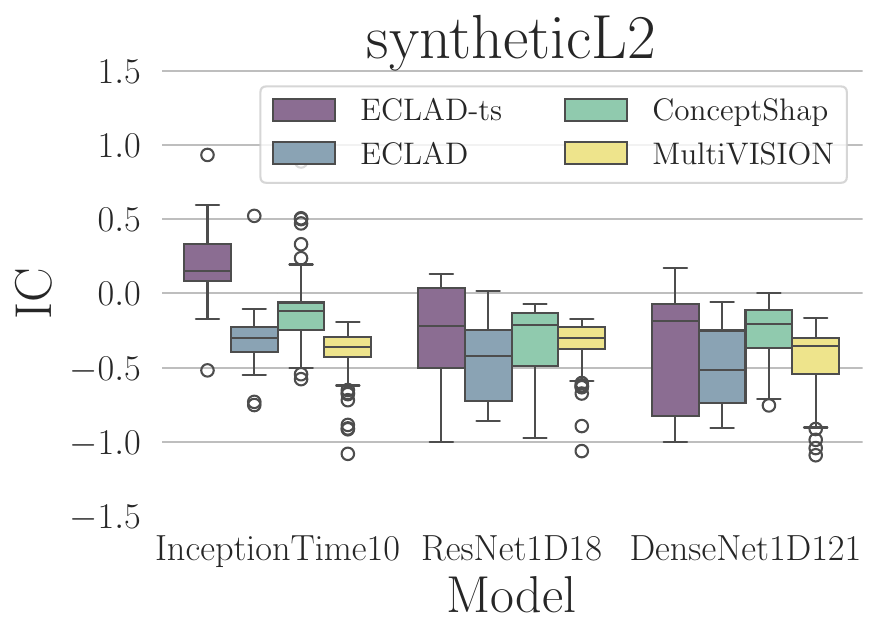}
        \end{subfigure}
    \end{minipage}
    \caption{
    Boxplots of (left) representation correctness and (right) importance correctness for CE methods on the syntheticL2 dataset.
    The box plots aggregate the data across all random seeds and the concept numbers chosen as a hyperparameter. All models achieved a validation accuracy of $100\%$.
    In both plots higher is better, as it means that the extracted concepts are more aligned with the ideal results of a concept extraction method. 
    The plots that are collapsed at $-0.200$ are methods that exclusively obtained the maximum penalty by failing to identify alignment. 
    ECLAD-based methods consistently achieve higher representation correctness, while ECLAD-ts attains the best importance correctness overall, highlighting its ability to extract concepts which are aligned with the underlying primitives.
    }
    \label{fig:L2_metrics}
\end{figure}

The representation and importance correctness metrics for all CE methods on the syntheticL2 dataset are shown in Fig.~\ref{fig:L2_metrics}. 
In terms of representation correctness, ECLAD-ts and ECLAD are, as expected, very similar.
In addition, they consistently outperform the other methods, showing a better alignment between the extracted concepts and the underlying primitives. 
Notably, due to the network's large receptive fields, MultiVISION is unable to localize any concepts beyond the background. 
Although ConceptShap localizes important concepts when used with InceptionTime, its performance is inconsistent across other models. 
Its high performance in that particular case is due to its masks often showing complete instances with one class, which can perfectly overlap with prmitive $p_1$ (uninterrupted background).  
Regarding importance correctness, the overall mean performance is similar across models; 
however, ECLAD-ts achieves the highest importance correctness compared to ECLAD, ConceptShap, and MultiVISION.
This shows that the modification to the IS in ECLAD-ts indeed improves its correctness. 
It is worth mentioning here that regardless of parameter tuning, ConceptShap failed to converge to coherent importance scores for several seeds and concept numbers.
This is partly due to the nature of ConceptShap's evaluation scheme, its Monte Carlo approximation, and the task at hand. 
The simple tasks in these datasets require only one meaningful concept for the model to succeed. 
ConceptShap always takes one concept out and re-trains its surrogate model, making the importance a function of the final accuracy difference. 
However, if the concepts are all redundant, a perfect accuracy is possible after taking each of the concepts out. 

\begin{figure}[!h]
    \begin{minipage}{.48\linewidth}
    \centering ECLAD-ts\\
        \begin{subfigure}[t]{\linewidth}
            \includegraphics[width=\linewidth]{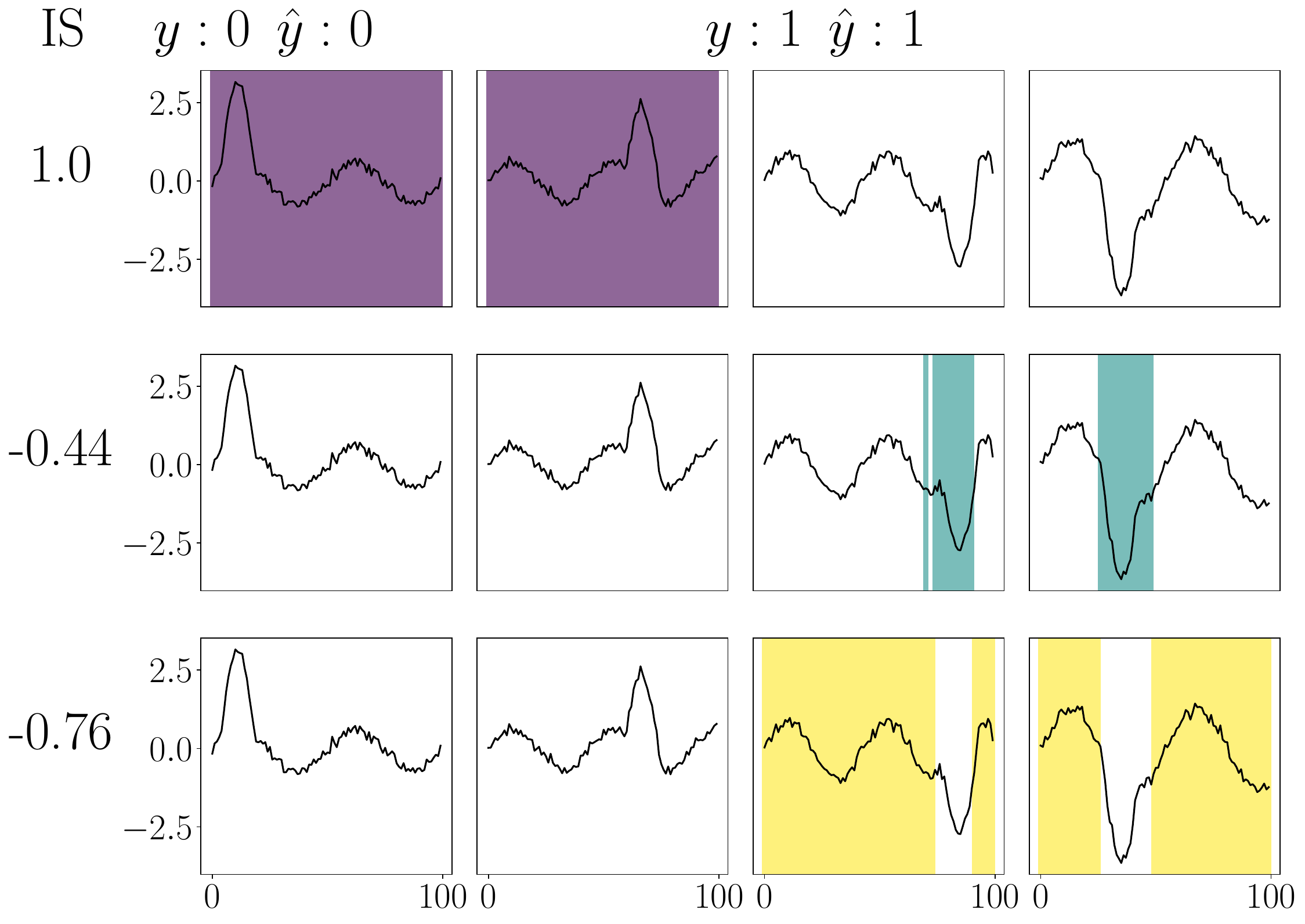}
        \end{subfigure}
    \end{minipage}%
    \begin{minipage}{.48\linewidth}
    \centering ECLAD\\
        \begin{subfigure}[t]{\linewidth}
            \includegraphics[width=\linewidth]{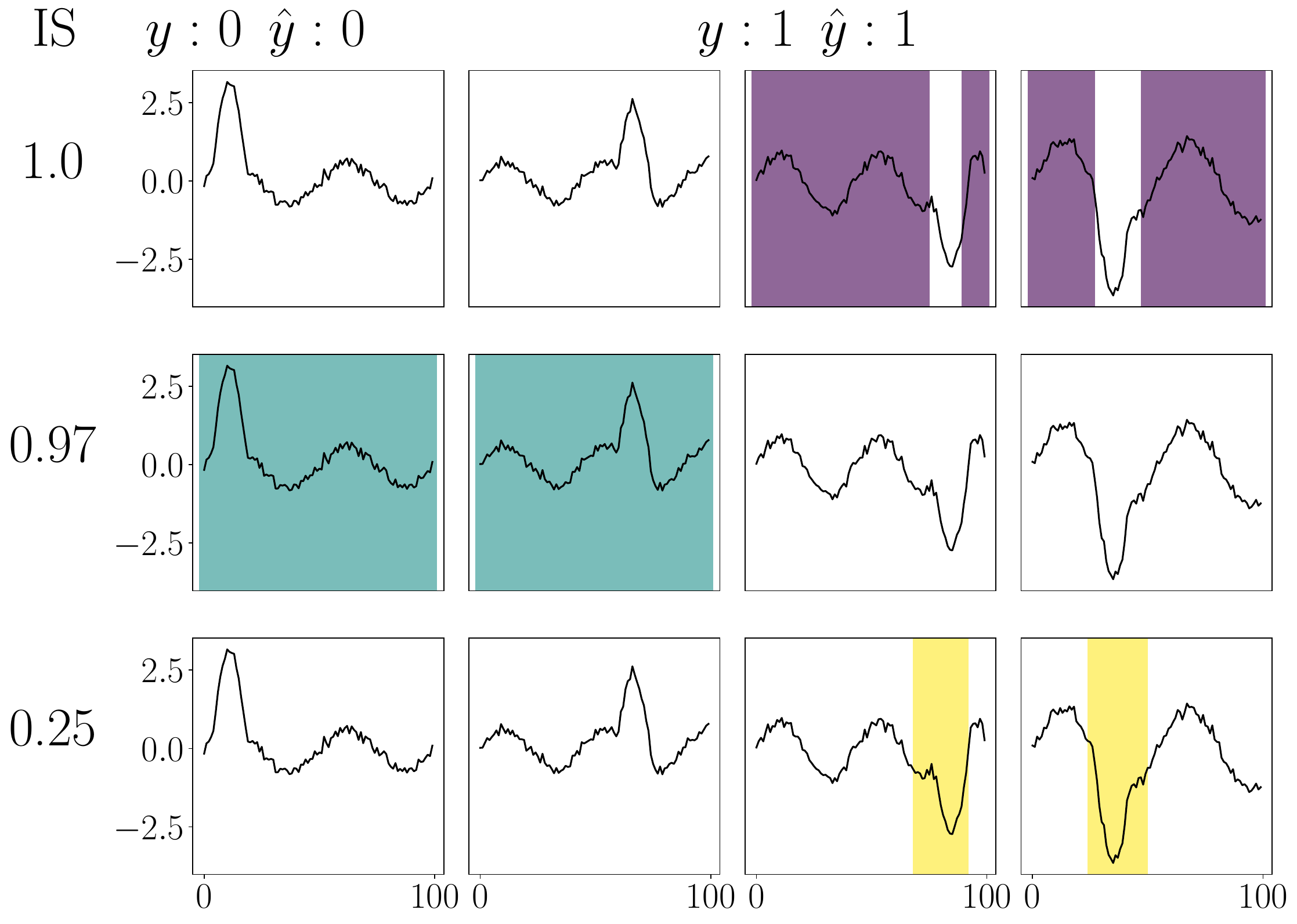}
        \end{subfigure}
    \end{minipage}
    \begin{minipage}{.48\linewidth}
    \centering ConceptShap\\
        \begin{subfigure}[t]{\linewidth}
            \includegraphics[width=\linewidth]{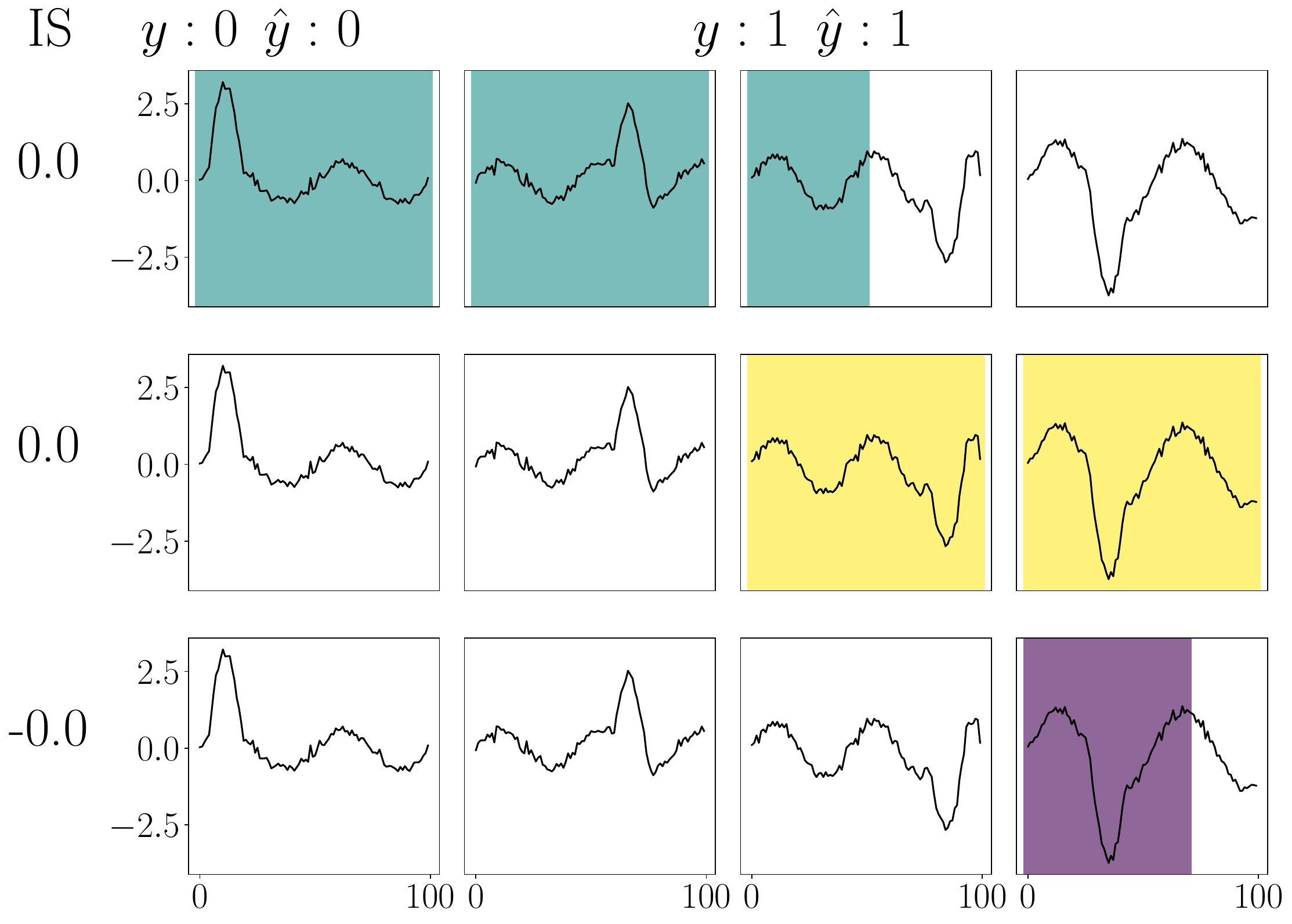}
        \end{subfigure}
    \end{minipage}%
    \begin{minipage}{.48\linewidth}
    \centering MultiVISION\\
        \begin{subfigure}[t]{\linewidth}
            \includegraphics[width=\linewidth]{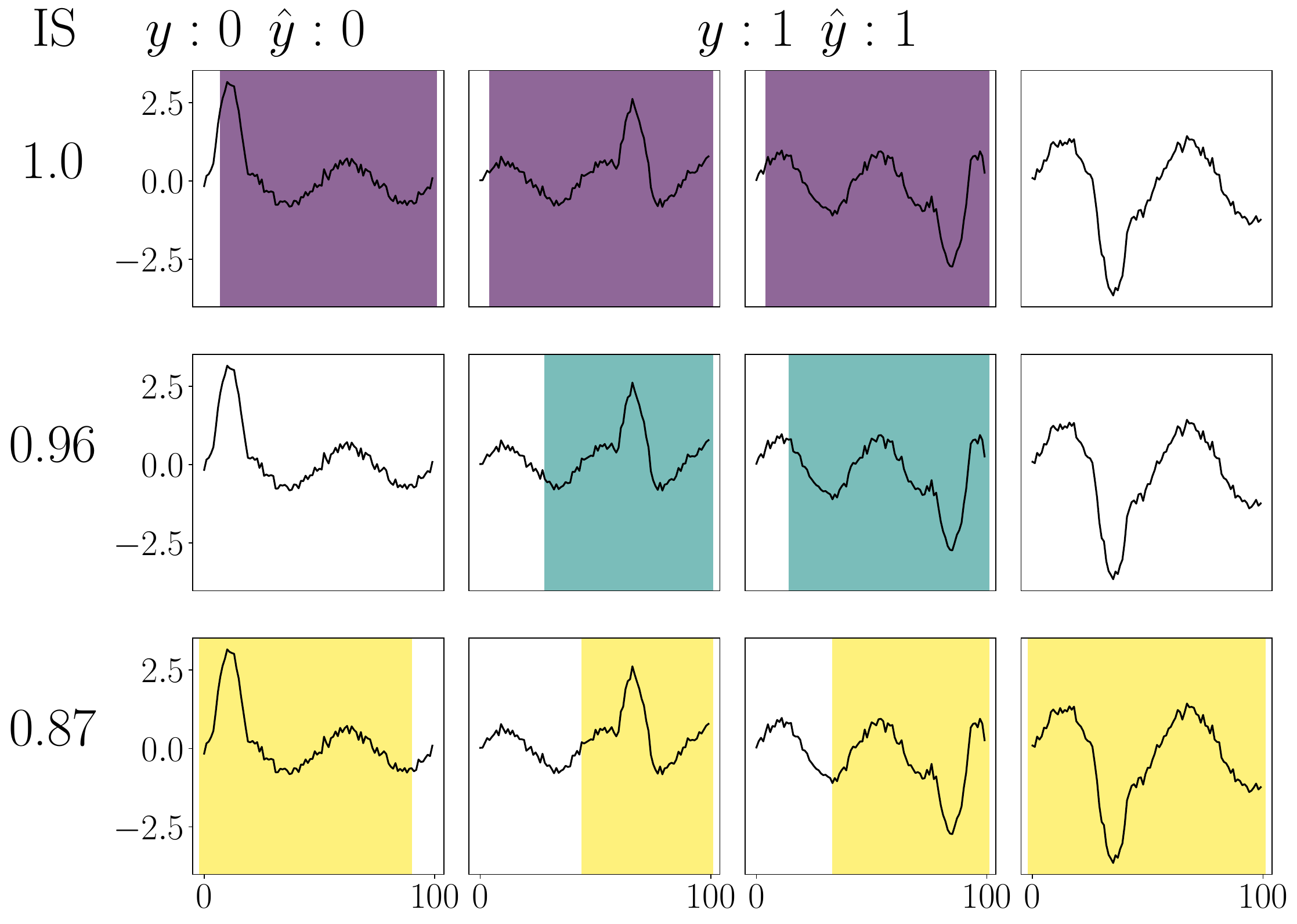}
        \end{subfigure}
        \end{minipage}
    \caption{
Concept extraction from InceptionTime on the SyntheticL4 dataset. 
The dataset is composed of a sinusoidal background with two primitives: an upward bump \(p_0\) for class 0 and a downward bump \(p_1\) for class 1. 
The model (seed=1) achieved a validation accuracy of $100\%$. 
In each panel, rows denote individual concepts and columns represent instances (with headers showing actual/predicted labels); 
highlighted regions indicate the presence of concepts, while left-hand labels report their importance scores. 
ECLAD-ts and ECLAD differentiate the \(p_1\) feature from the background, whereas ConceptShap and MultiVISION fail to capture coherent primitive-related features. 
    }
    \label{fig:L4_IT}
\end{figure}

The \textbf{SyntheticL4} dataset has a sinusoidal background and two primitives: a bump upwards $p_0$ for class 0 and a bump downwards $p_1$ for class 1.
Fig.~\ref{fig:L4_IT} presents the concepts extracted by each method from an InceptionTime model trained on this dataset. 
The concepts show that the model learns to recognize the backgrounds lacking $p_1$. 
The latent representations in ECLAD-ts and ECLAD do differentiate the bump down, as we can see in the other concepts.
However, according to ECLAD-ts' importance scores, this is also a case of shortcut-learning, where $p_1$ is primarily relied on for predictions. 
The distinction between background and features from the primitives is depicted by both ECLAD and ECLAD-ts, but not by ConceptShap and MultiVISION. 
Both ConceptShap and MultiVISION fail to extract coherent features from the model.

\begin{figure}[!ht]
    \begin{minipage}{.5\linewidth}
        \begin{subfigure}[t]{\linewidth}
            \includegraphics[width=\linewidth]{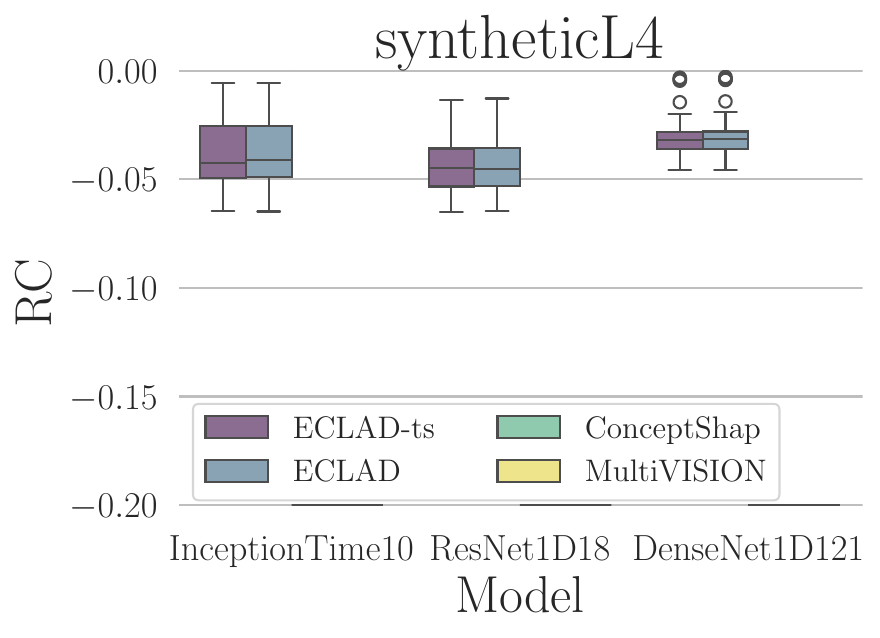}
        \end{subfigure}
    \end{minipage}%
    \begin{minipage}{.5\linewidth}
        \begin{subfigure}[t]{\linewidth}
            \includegraphics[width=\linewidth]{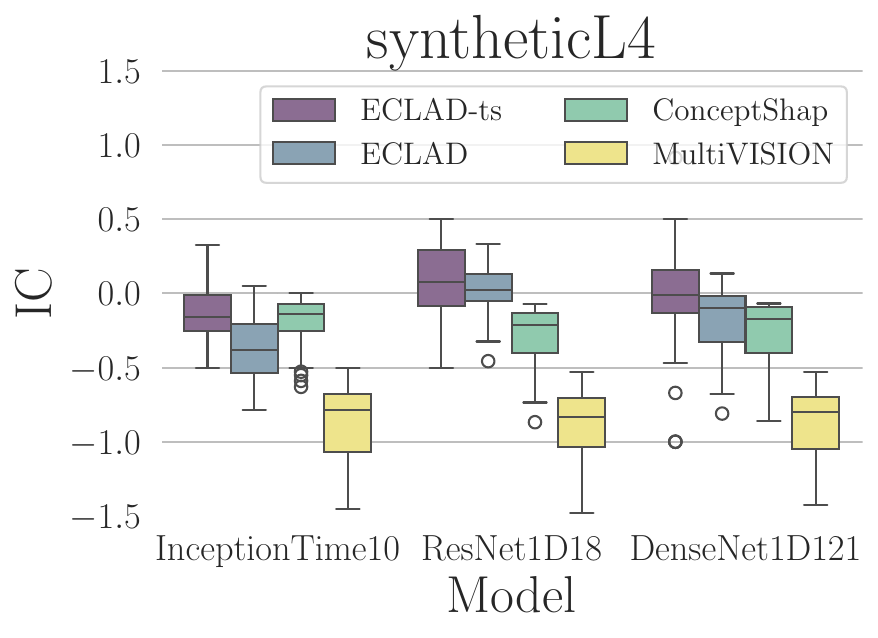}
        \end{subfigure}
        \end{minipage}
    \caption{
Boxplots of (left) representation correctness and (right) importance correctness for CE methods on the SyntheticL4 dataset.
The box plots aggregate the data across all random seeds and the concept numbers.
All models achieved a validation accuracy of $100\%$.
Higher scores indicate better alignment of extracted concepts with the underlying primitives. 
The plots that are collapsed at $-0.200$ are methods that exclusively obtained the maximum penalty by failing to identify alignment. 
ECLAD-based methods consistently outperform ConceptShap and MultiVISION, with the latter methods scoring -0.2 in representation correctness—highlighting their inability to localize primitive features—while ECLAD-ts achieves the best importance correctness overall.
    }
    \label{fig:L4_metrics}
\end{figure}

The metrics in Fig.~\ref{fig:L4_metrics} show, similar as before, that the representation correctness of the ECLAD-based methods is better than for the other methods. 
ConceptShap and MultiVISION always get a representation correctness score of -0.2, which corresponds to the 40\% of the maximum penalty given when no alignment is detected, indicating that they are not able to localize the features in the primitives. 
In the case of the importance correctness,  we still see increased performance from ECLAD-ts \wrt the other methods. 

The third synthetic dataset, \textbf{syntheticLm}, contains two channels and two primitives. 
Primitive \(p_0\) is located in the first channel and \(p_1\) in the second channel. 
The presence of \(p_0\) determines class 0, the presence of \(p_1\) determines class 1, and class 2 occurs when neither primitive is present; they never co-occur. 
This dataset highlights a key challenge in explaining time series: features can be distinctly encoded across channels, making channel-wise localization critical. 
Thus, an ideal CE method should be capable of isolating and localizing such features.

\begin{figure}[htbp]
    \begin{minipage}{0.1\linewidth}
         \rotatebox{90}{\scriptsize ECLAD-ts}
    \end{minipage}
    \begin{minipage}{0.9\linewidth}
        \begin{subfigure}[t]{\linewidth}
            \includegraphics[width=\linewidth]{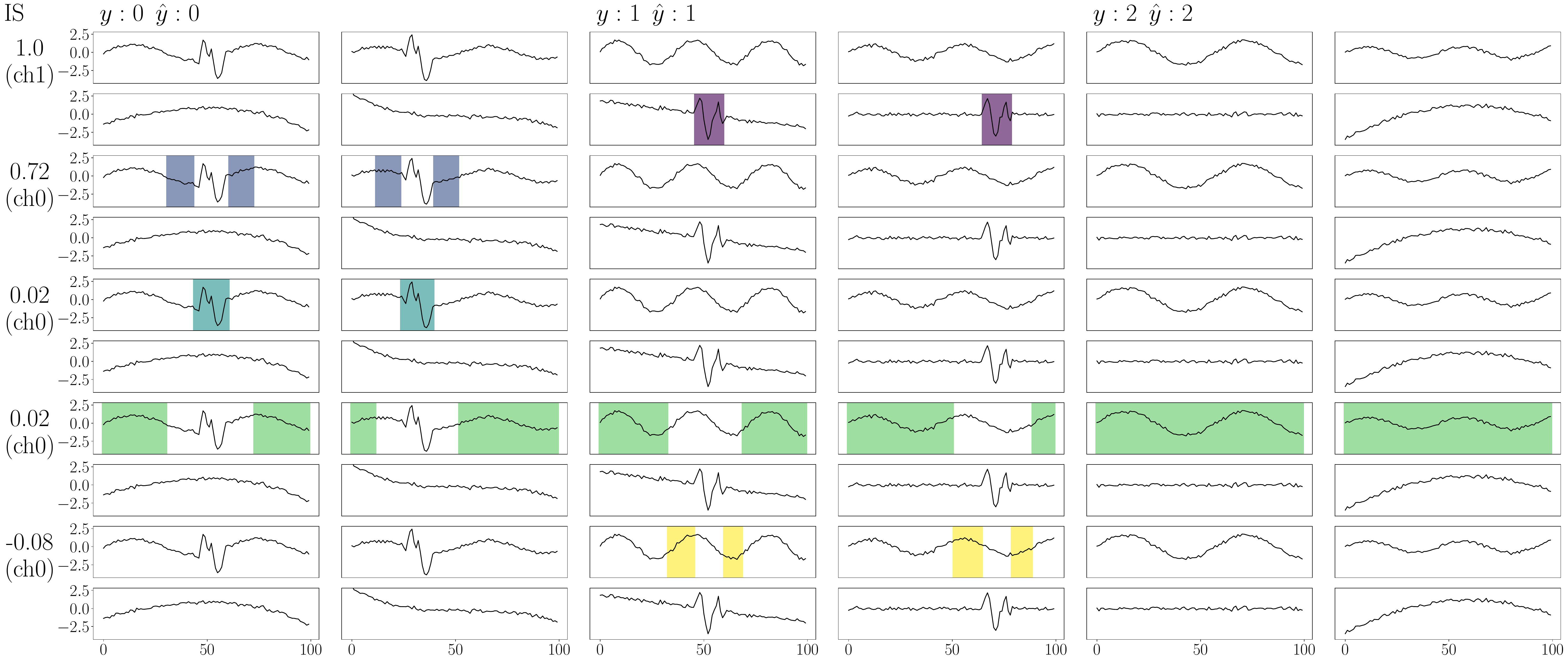}
        \end{subfigure}
    \end{minipage}
    
    \begin{minipage}{0.1\linewidth}
         \rotatebox{90}{\scriptsize ECLAD}
    \end{minipage}
    \begin{minipage}{0.9\linewidth}
        \begin{subfigure}[t]{\linewidth}
            \includegraphics[width=\linewidth]{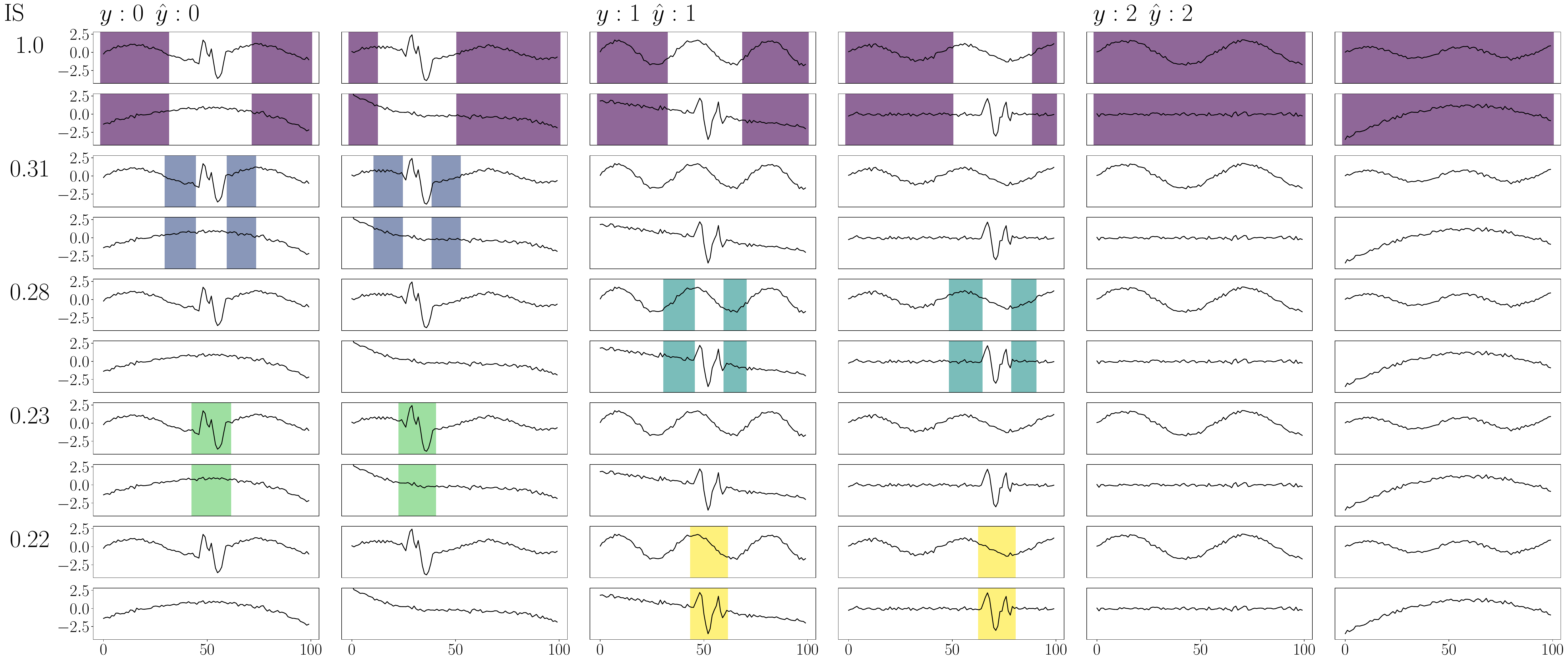}
        \end{subfigure}
    \end{minipage}
    
    \begin{minipage}{0.1\linewidth}
         \rotatebox{90}{\scriptsize ConceptShap}
    \end{minipage}
    \begin{minipage}{0.9\linewidth}
        \begin{subfigure}[t]{\linewidth}
            \includegraphics[width=\linewidth]{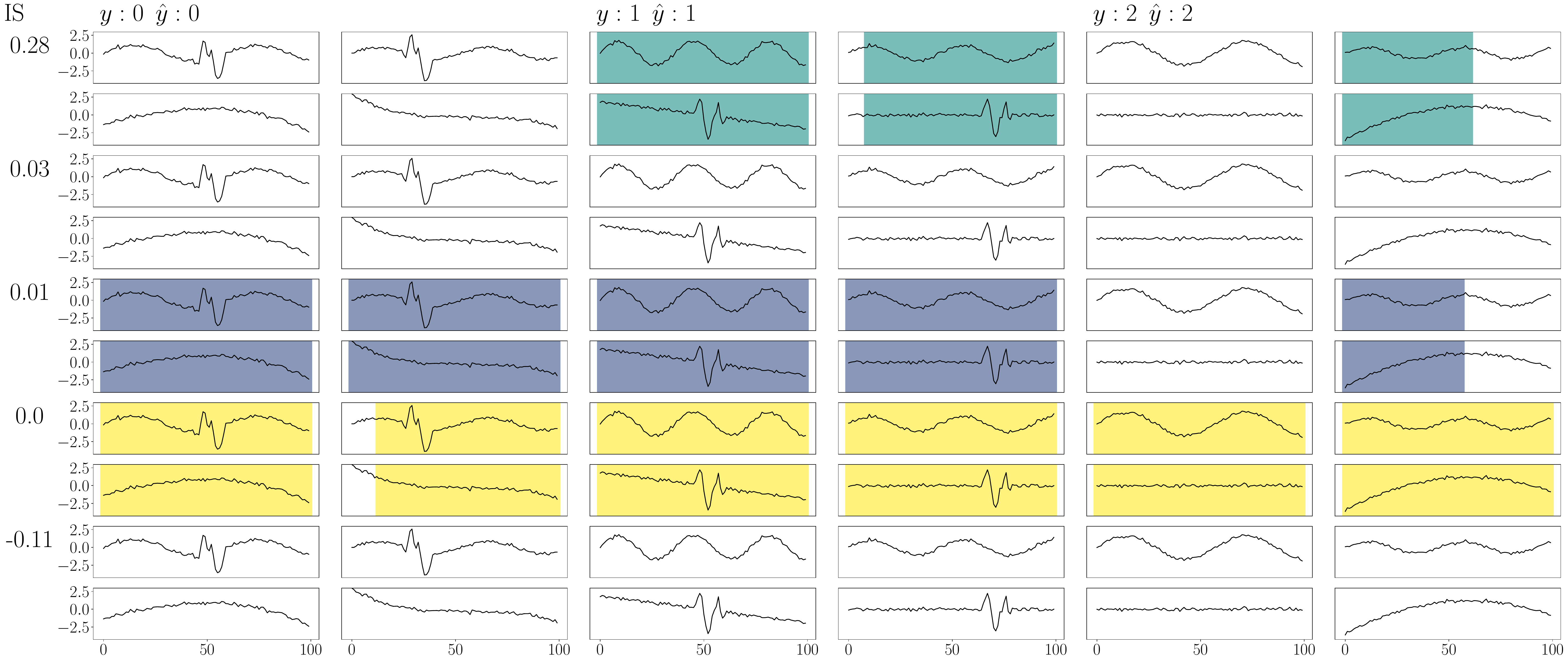}
        \end{subfigure}
    \end{minipage}
    
    \begin{minipage}{0.1\linewidth}
         \rotatebox{90}{\scriptsize MultiVISION}
    \end{minipage}
    \begin{minipage}{0.9\linewidth}
        \begin{subfigure}[t]{\linewidth}
            \includegraphics[width=\linewidth]{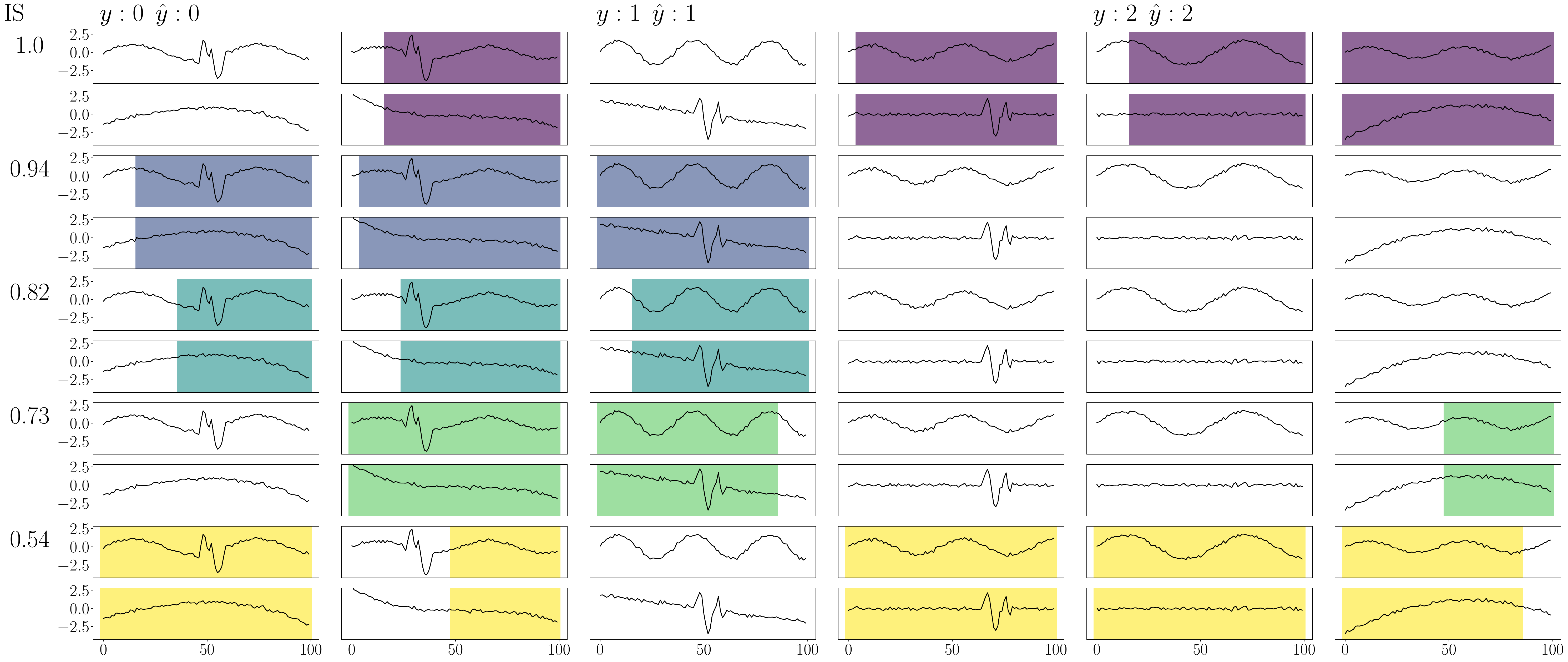}
        \end{subfigure}
        \end{minipage}
    \caption{
Concept extraction from a DenseNet121 (seed=1) with $100\%$ validation accuracy trained on the syntheticLm dataset. In each panel, a pair of rows (channels ) represents a concept and each column a sample, with headers showing ground truth/predicted labels and left labels indicating the importance scores. For concepts sharing a centroid in ECLAD-ts, only the most important concept is shown, and its channel index is indicated as `ch'. 
Notably, ECLAD-ts localizes \(p_1\) in concept 0 and \(p_0\) in concept 3, whereas ECLAD fails to differentiate channels and both ConceptShap and MultiVISION yield ambiguous results.
    }
    \label{fig:LM_DN}
\end{figure}

In Fig.~\ref{fig:LM_DN}, the concepts extracted by each of the methods are shown consecutively. 
For ECLAD-ts, we can see that concepts 0 and 3 correspond clearly to primitives $p_1$ and $p_0$, respectively. 
ECLAD is able to localize the same concepts through time, but not able to differentiate them through channels. 
In contrast, it is unclear what the extracted concepts from ConceptShap and MultiVISION refer to.
The particularity of time series containing different information in each channel is not taken into account by ECLAD, ConceptShap or MultiVISION. 
With the extracted concepts of the ECLAD-based methods, we can analyze the inputs in the columns of Fig.~\ref{fig:L4_IT} and understand how they are classified according to the concepts present in them. For example, for the sample in the first column, using ECLAD-ts: InceptionTime classifies it as class 0 because it detects $p_0$, and gives little importance to the background. 

\begin{figure}[!h]
    \centering
    \begin{minipage}{.5\linewidth}
        \begin{subfigure}[t]{\linewidth}
            \includegraphics[width=\linewidth]{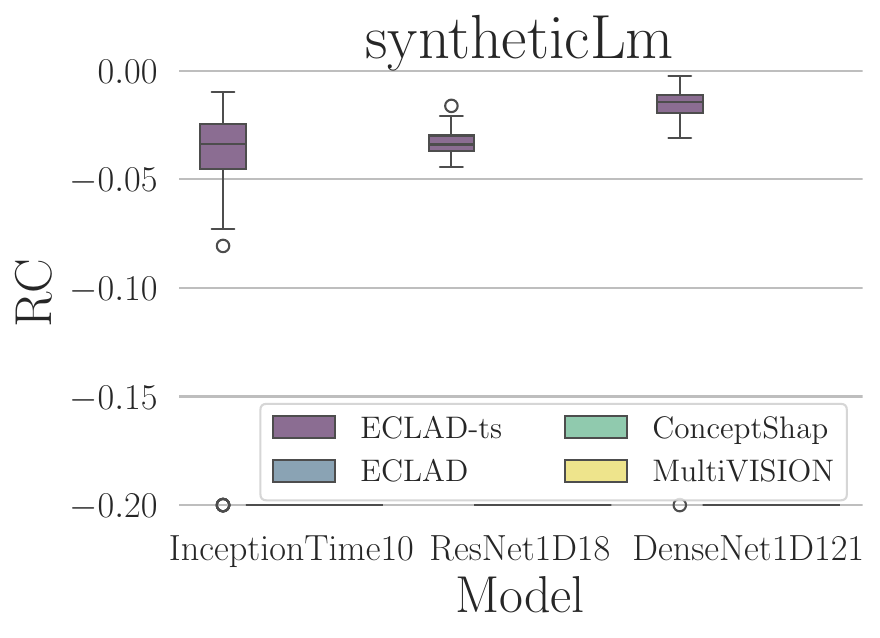}
        \end{subfigure}
    \end{minipage}%
    \begin{minipage}{.5\linewidth}
        \begin{subfigure}[t]{\linewidth}
            \includegraphics[width=\linewidth]{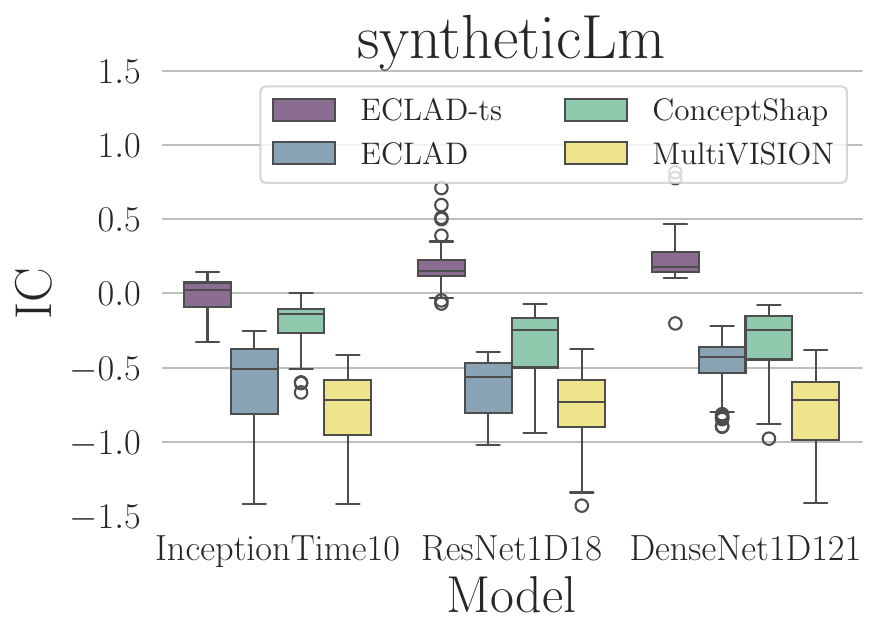}
        \end{subfigure}
    \end{minipage}
    \caption{
Boxplots of (left) representation correctness and (right) importance correctness for CE methods on the syntheticLm dataset.
The box plots aggregate the data across all random seeds and the concept numbers chosen as a hyperparameter.
All models achieved a validation accuracy of $100\%$.
Higher scores indicate better alignment of extracted concepts with the underlying channel-specific primitives. The plots that are collapsed at $-0.200$ are methods that exclusively obtained the maximum penalty by failing to identify alignment. 
ECLAD-ts is the only method that achieves representation alignment and obtains the highest mean importance correctness.
    }
    \label{fig:LM_metrics}
\end{figure}

The capability of ECLAD-ts to localize features channel-wise is translated into the representation and importance correctness metrics, as shown in Fig.~\ref{fig:LM_metrics}.
We observe that ECLAD-ts is the only method that achieves concept alignment, and that it has a better mean importance correctness than all other methods. 
This shows that the proposed method is able to recognize channel-wise features encoded differently and score their importance accordingly, better than other methods. 

CE with synthetic datasets demonstrates that models \textbf{encode input patterns into latent space patterns}, and that these patterns and their importance for the prediction process can be identified using ECLAD-ts.
Additionally, we confirm that \textbf{ECLAD-ts performs as desired}: The concepts directly highlight the primitives, localizing them in time and channel-wise, the distance between concepts and their primitives is small, and the importance correctness achieved is higher than for the compared methods.  %
We also show that \textbf{extracted concepts provide insights on the model}, like the most important features for the model or the presence of shortcut learning. 
These insights allow us to assess the alignment of the model with human expectations and make informed decisions with respect to the model to avoid unnecessary risks. 

\subsubsection{Evaluation on Natural Datasets}
\begin{figure}[!h]
    \begin{minipage}{.48\linewidth}
    \centering ECLAD-ts\\
        \begin{subfigure}[t]{\linewidth}
            \includegraphics[width=\linewidth]{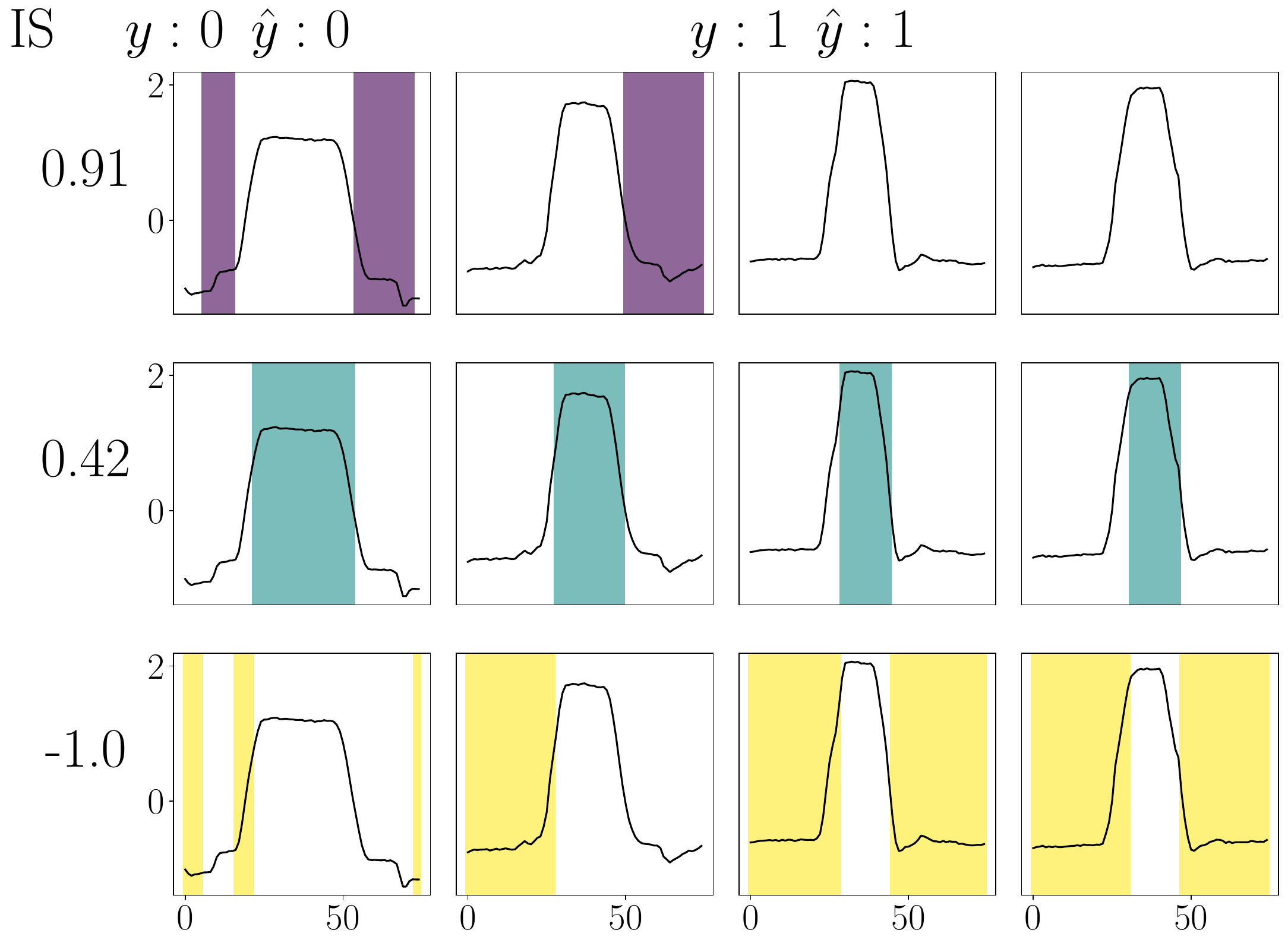}
        \end{subfigure}
    \end{minipage}%
    \begin{minipage}{.48\linewidth}
    \centering ECLAD\\
        \begin{subfigure}[t]{\linewidth}
            \includegraphics[width=\linewidth]{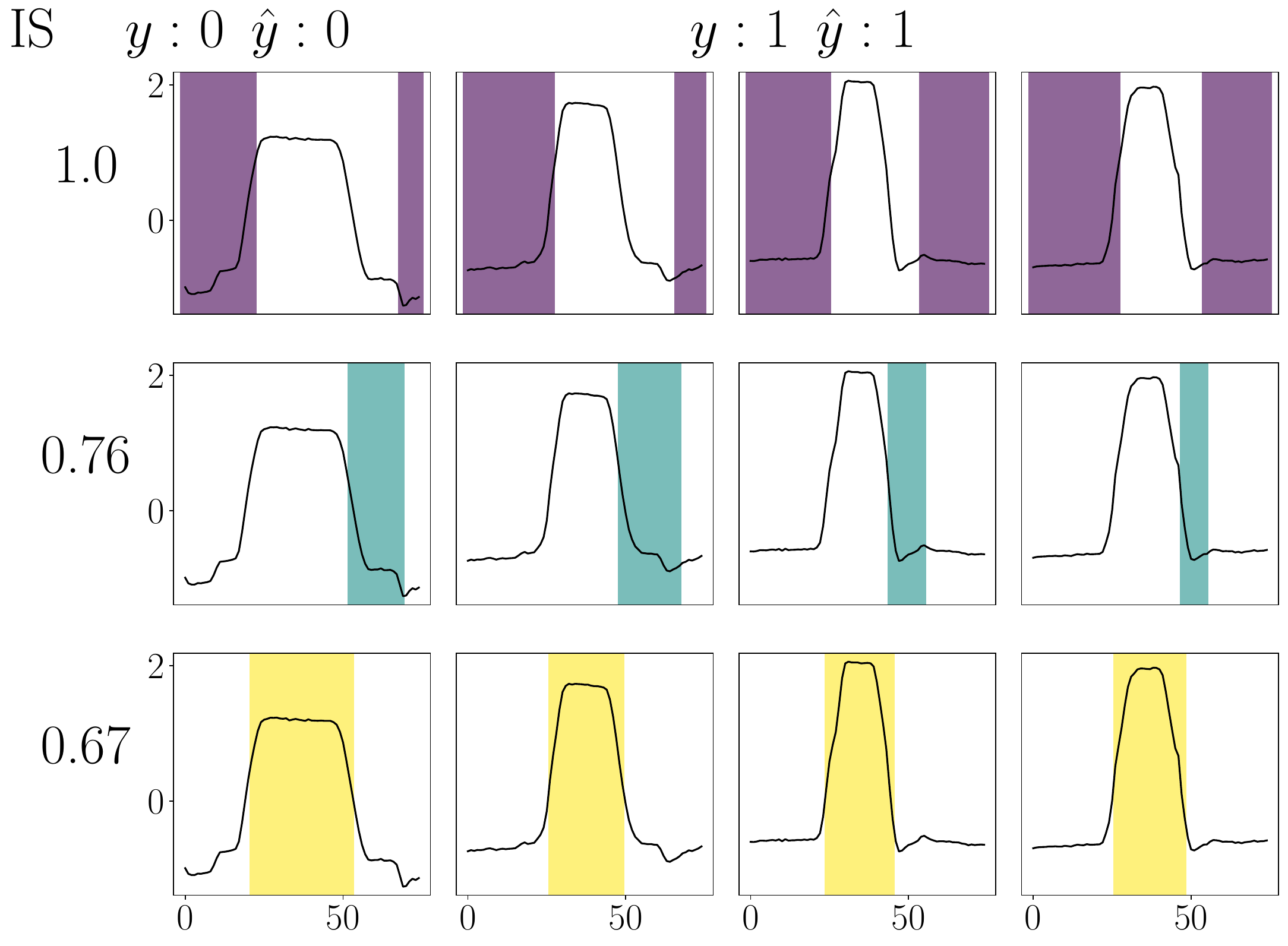}
        \end{subfigure}
    \end{minipage}
    \begin{minipage}{.48\linewidth}
    \centering ConceptShap\\
        \begin{subfigure}[t]{\linewidth}
            \includegraphics[width=\linewidth]{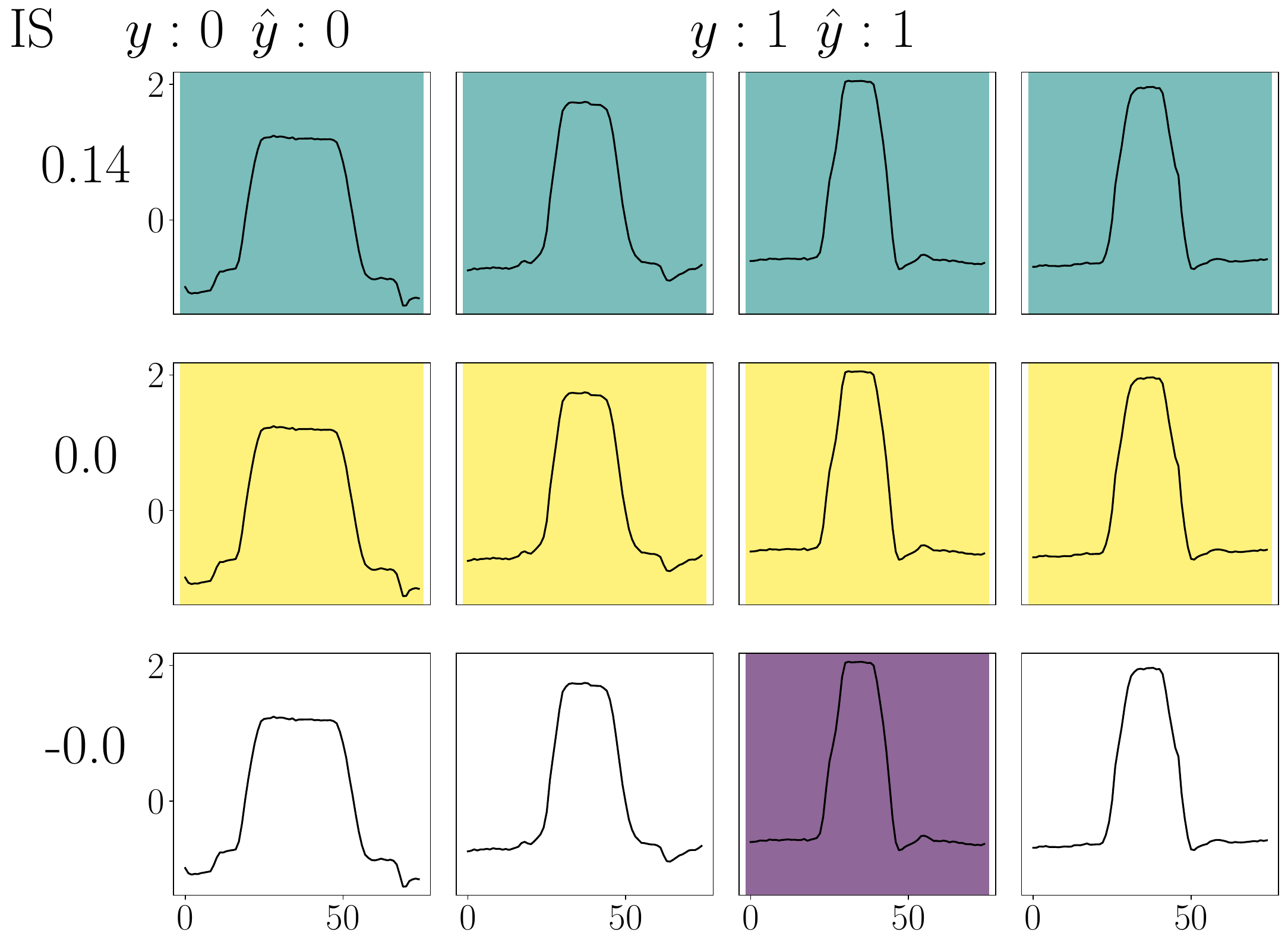}
        \end{subfigure}
    \end{minipage}%
    \begin{minipage}{.48\linewidth}
    \centering MultiVISION\\
        \begin{subfigure}[t]{\linewidth}
            \includegraphics[width=\linewidth]{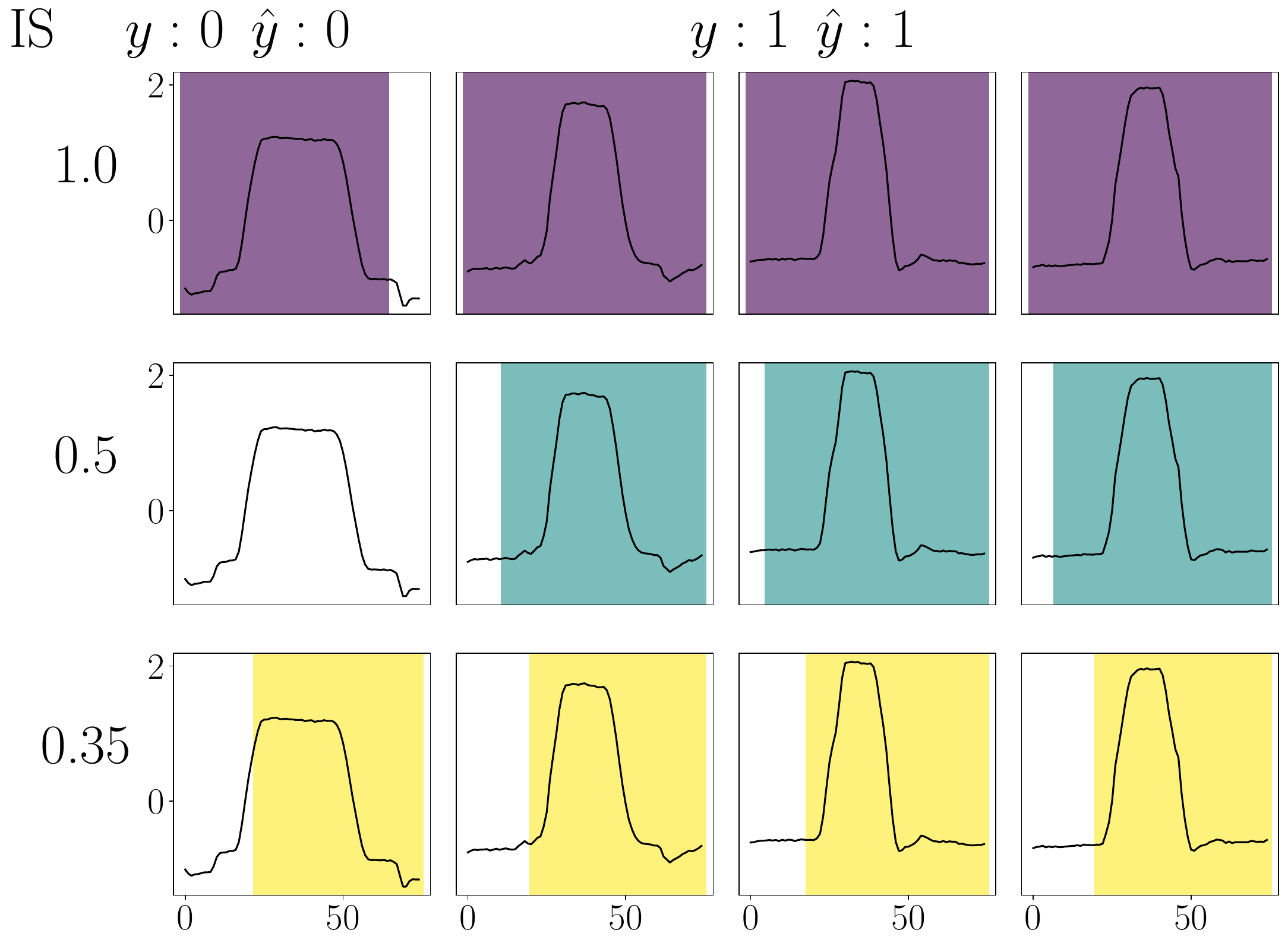}
        \end{subfigure}
        \end{minipage}
    \caption{Concept extraction from InceptionTime on the GunPoint dataset is depicted for four methods. The model (seed=1) achieves a $96.6\%$ validation accuracy.
            The dataset contains sensor data on hand movement for actors holding a prop gun or not. 
            Two parts of the input are known to be the most significant: the `extra lifting' movement before the higher position and the `overshoot' motion when arriving at the lower position~\cite{hills2014classification}. 
            In each panel, rows denote individual concepts and columns represent instances (with headers showing actual/predicted labels: 
            the first two for class 0 and the latter two for class 1), highlighted regions indicate where concepts appear, and left labels report their importance scores. 
            ECLAD-ts and ECLAD extract concepts related to the specified features (`overshoot' and `extra lifting' motions), while ConceptShap and MultiVISION capture background cues.}
    \label{fig:GunPoint}
\end{figure}

We now examine the results of CE with InceptionTime models trained on the \textbf{GunPoint} dataset, shown in Fig.~\ref{fig:GunPoint}. The dataset has position sensor data for the hand movement of actors holding a prop gun (class 0) or not (class 1) while making the motions of drawing, pointing and then lowering the prop. The dataset is included due to it having real world sensor data where it is known which specific features are important for the prediction~\cite{hills2014classification}. These correspond to the ‘overshoot’ motion 
when lowering the arm in class 1 and the `extra lifting' movement 
in class 0 (or the lack of such features in the opposite class). 

For the InceptionTime model, we observe that the ECLAD-based methods both observe the important concepts corresponding to the `overshoot' and `extra lifting' motions. %
These insights allow us to asses the model alignment with human expectations. 
In contrast to this, ConceptShap and MultiVISION produce very broad concept masks that are inconclusive \wrt the the important features. 


Finally, we test the CE methods using the \textbf{P2S} dataset, which shows their performance in an industrial setting.
The Production Press Sensor Data (P2S) dataset consists of sensor recordings from a sheet metal production process involving stamping, deep drawing, and bending operations. 
Each sample is a time series capturing internal forces, with the goal of classifying production runs as normal or faulty. 

\begin{figure}[!h]
    \begin{minipage}{.5\linewidth}
    \centering ECLAD-ts\\
        \begin{subfigure}[t]{\linewidth}
            \includegraphics[width=\linewidth, height=0.6\linewidth]{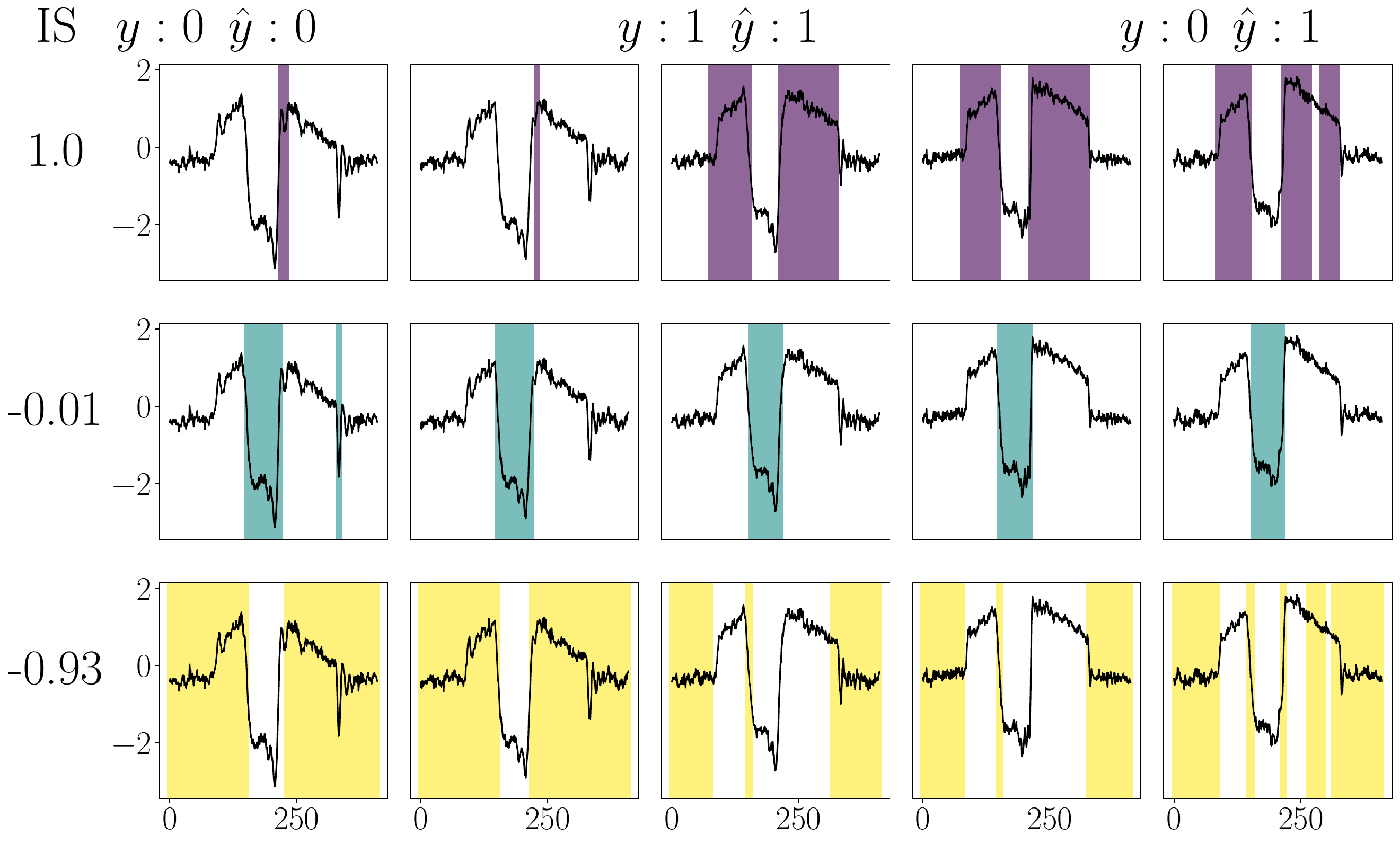}
        \end{subfigure}
    \end{minipage}%
    \begin{minipage}{.5\linewidth}
    \centering ECLAD\\
        \begin{subfigure}[t]{\linewidth}
            \includegraphics[width=\linewidth, height=0.6\linewidth]{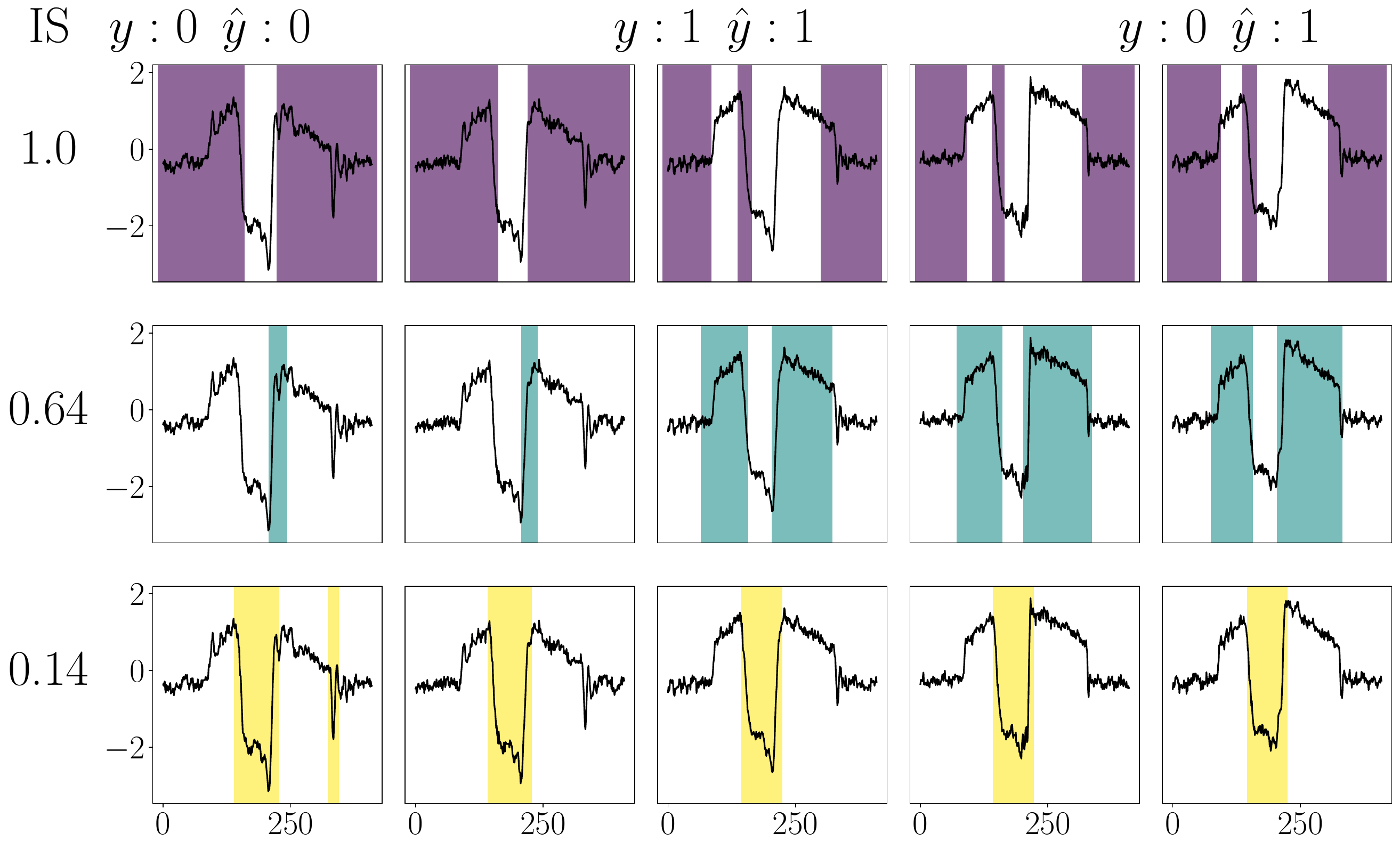}
        \end{subfigure}
    \end{minipage}
    \begin{minipage}{.5\linewidth}
    \centering ConceptShap\\
        \begin{subfigure}[t]{\linewidth}
            \includegraphics[width=\linewidth, height=0.6\linewidth]{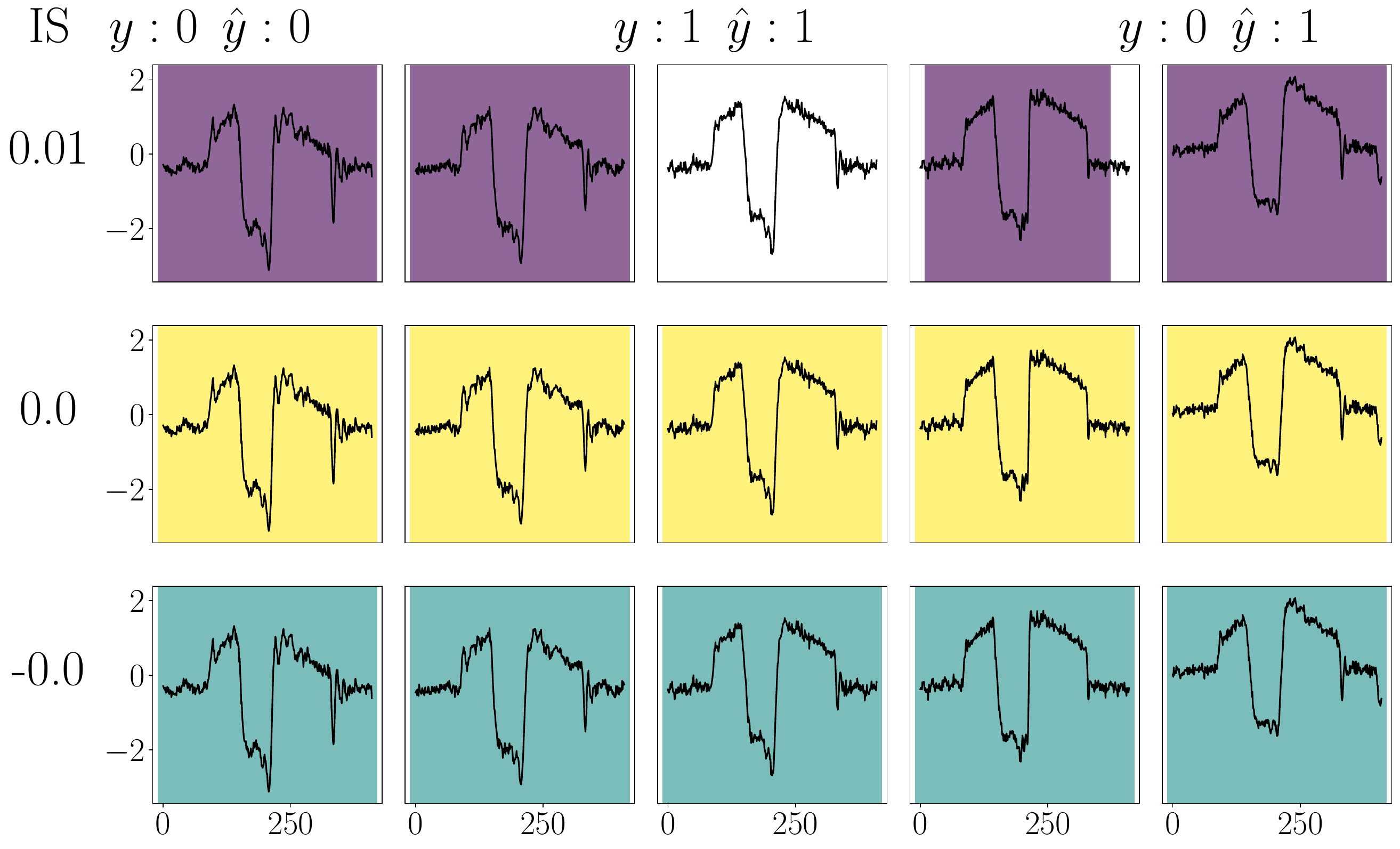}
        \end{subfigure}
    \end{minipage}%
    \begin{minipage}{.5\linewidth}
    \centering MultiVISION\\
        \begin{subfigure}[t]{\linewidth}
            \includegraphics[width=\linewidth, height=0.6\linewidth]{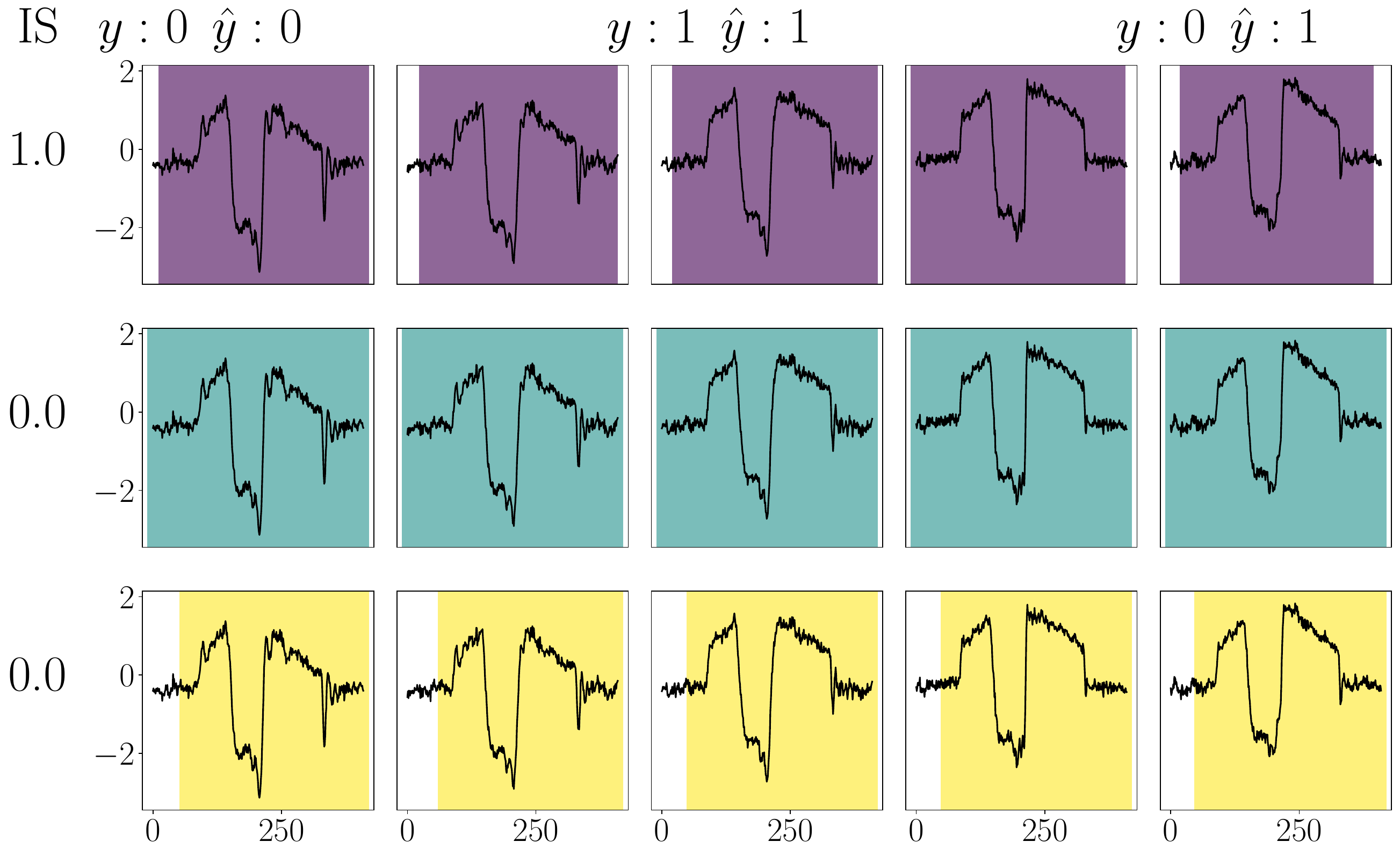}
        \end{subfigure}
        \end{minipage}
    \caption{Concept extraction from InceptionTime on the P2S dataset is illustrated for four methods. The model has seed 0 and achieved a validation accuracy of $99.7\%$. 
            The dataset consists of sensor recordings of a part production process at different speeds, with either normal or faulty outcomes. 
            The most significant part of the input according to experts is the area are the beginning and end of the slopes before the concavity in the middle. 
            In each panel, rows denote individual concepts and columns represent instances (with headers showing actual/predicted labels: 
            the first two for class 0, the latter two for class 1), and the final column referring to misclassified instances.
            Highlighted regions indicate where concepts appear, and left labels report their importance scores. 
            ECLAD-ts and ECLAD extract concepts related to the aforementioned features, while ConceptShap and MultiVISION capture background cues.}
    \label{fig:P2S}
\end{figure}

CE on the P2S dataset in Fig.~\ref{fig:P2S} offers similar insights as those in the GunPoint dataset. 
The datasets' most important features are located on the slopes before the concavity in the middle. 
ECLAD-ts and ECLAD both produce concepts that distinguish between the background, slopes of one type, and the concavity. 
Furthermore, ECLAD-ts assigns a very high importance to the slopes of class 1.
We can additionally examine the missclassified sample on the right of Fig.~\ref{fig:P2S}.
It shows us that the model has mistaken the slope of a class 0 sample with the one of class 1. 
Observing the concepts present in missclassified samples (and \eg further examining the confounding slope) can be a good insight for practitioners, and help improve the models. 

From the CE on natural datasets GunPoint and P2S models, we see that ECLAD-ts is capable of producing \textbf{explanations for models trained on real datasets} with relevant local structures,\textbf{ revealing biases and shortcut learning}. The obtained insights allow humans to understand which patterns the model is using \textbf{generally and for single instances}, why a model makes a particular prediction and which risks exist in the prediction process.

Overall, our results demonstrate that CNNs encode meaningful latent patterns which can be effectively extracted using our concept extraction approach.
The experiments on both synthetic and natural datasets showed that ECLAD-based methods have a robust capability for localizing relevant features.
In particular, ECLAD-ts can do this in the channel dimension, which is crucial for explaining time series models. 
Additionally, ECLAD-ts consistently outperforms other concept extraction methods (ECLAD, ConceptShap, and MultiVISION) across both synthetic and natural datasets, in the granularity of the obtained masks, and in the representation and importance correctness metrics.
Given this, CE with ECLAD-ts provides significant insights into model behavior for \eg detecting shortcut learning and understanding the most important factors for a prediction.

\section{Conclusion}

We propose ECLAD-ts as the first global automatic concept extraction and localization method specifically designed for time series. ECLAD-ts addresses the unique challenges of time series data—particularly its high dimensionality and the distinct functionalities of different channels—by extracting and localizing concepts not only along the temporal axis but also across channels.

In our work, we introduce three key contributions.
First, we introduce a novel method that leverages Local Aggregated Descriptors (LADs) to extract meaningful patterns from the latent space of CNNs, capturing both temporal and channel-specific information. 
Second, we develop an importance scoring mechanism that quantitatively assesses the relevance of a concept in a channel-wise manner.
Third, through extensive experiments on synthetic and real-world datasets, we demonstrate that ECLAD-ts outperforms existing methods (ECLAD, ConceptShap, and MultiVISION) in terms of representation and importance correctness—extracting concepts that are more precisely aligned with ground truth primitives both temporally and channel-wise.

In comparison with other methods, which rely on shapley values, ECLAD-ts is more efficient and can be easily integrated for explaining CNNs, making it scalable and suitable for large-scale, real-world applications. 
While our experiments focus on synthetic data and industrial classification, the underlying framework is broadly applicable to any field involving multivariate time series, such as predictive maintenance, financial forecasting, environmental monitoring, and sensor data analysis.

In our current work, multiple challenges were observed. In particular, the visualization of complex features in the time series domain can be problematic.
Similarly, all considered concept extraction methods exhibit sensitivity to hyperparameter settings, especially to the number of concepts, which highlights the importance of developing more adaptive techniques.
Moreover, concept compositionality and correlation, which is a common phenomenon in time series, can also be explored in the context of CE.
Future work will focus on refining these aspects and exploring the integration of concept extraction into real-time model training.

Here, we have focused on porposing ECLAD-ts, yet it is still an open question whether other methods like ConceptShap and MultiVISION can be modified to work properly in the time series domain. 
For these two methods, this could be achieved, \eg by producing visualizations based on a meaningful subset of the receptive field. 

In summary, ECLAD-ts represents a step toward better understanding the decision-making processes of convolutional neural networks for time series classifiers, enhancing model transparency, which can improve trust across diverse application domains.

\begin{credits}
\subsubsection{\ackname} We would like to thank Alexander Gräfe for his support in the upscaling of the experiments in this work. 
In addition, model training and experiments were partially performed with computing resources granted by RWTH Aachen University, under project p0022034.

\subsubsection{\discintname}
This work is partially funded by the Deutsche Forschungsgemeinschaft (DFG) within the Priority Program SPP 2422 (TR 1433/3-1) and Germany's Excellence Strategy--EXC-2023 Internet of Production -- 390621612.
\end{credits}
%
%
%
%

\bibliography{refs}
\end{document}